%% file: MDM_arxiv.tex
\documentclass{article}

\usepackage [nonatbib,preprint]{neurips_2021}

\usepackage{mathrsfs}

\usepackage[numbers,sort,compress]{natbib}

\include{arxiv/header}

\title{Marginalizable Density Models}

\author{
   Dar Gilboa \\
   Harvard University \\
   \texttt{dar\_gilboa@fas.harvard.edu} \\
   \And
   Ari Pakman \\
   Columbia University \\
   \texttt{ari@stat.columbia.edu} \\
   \And
   Thibault Vatter \\
   Columbia University \\
   \texttt{thibault.vatter@columbia.edu} \\
}

\begin{document}
\maketitle
\begin{abstract}
Probability density models based on deep networks have achieved
remarkable success in modeling complex high-dimensional datasets.
However, unlike kernel density estimators, modern neural models do not yield marginals or conditionals in closed form, 
as these quantities require the evaluation of seldom tractable integrals.
In this work, we present the \emph{marginalizable density model approximator} (MDMA),
a novel deep network architecture which provides closed form expressions for the probabilities, marginals and conditionals of any subset of the variables.
The MDMA learns deep
scalar representations for each individual variable
and combines them via learned hierarchical tensor decompositions into a tractable yet expressive CDF,
from which marginals and conditional densities are easily obtained.
We illustrate the advantage of exact marginalizability in several tasks that are out of reach of previous deep network-based density estimation models, such as estimating mutual information between arbitrary subsets of variables, inferring causality by testing for conditional independence, and inference with missing data without the need for data imputation, outperforming state-of-the-art models on these tasks. The model also allows for parallelized sampling with only a logarithmic dependence of the time complexity on the number of variables.  
\end{abstract}

\input{arxiv/intro.tex}

\input{arxiv/litterature}
\input{arxiv/mdma2}

\input{arxiv/theory}
\section{Experiments}
Additional experimental details for all experiments are provided in \Cref{app:exp_details}.\footnote{Code for reproducing all experiments is available at \url{https://github.com/dargilboa/mdma}.}
\label{sec:experiments}
\input{arxiv/exp_toy_combined}
\input{arxiv/exp_mie}
\input{arxiv/exp_demv}
\input{arxiv/exp_causal}
\input{arxiv/exp_UCI}
\input{arxiv/discussion}

\newpage 

\input{arxiv/ack}
\bibliographystyle{unsrt}  
\bibliography{thebib}

\newpage

\appendix
{\Large \bf{Supplementary Material}}
\input{arxiv/proofs}
\section{Additional experimental results} \label{app:additional_exps}
\subsection{Toy density estimation}

Figures \ref{fig:checkerboard} and \ref{fig:8gaussians}
show more results on the popular checkerboard and 8 Gaussians toy datasets
studied in \Cref{fig:toy_de}. 
\input{arxiv/exp_toy_checkerboard}
\input{arxiv/exp_toy_gaussians}

\subsection{Density estimation with missing data}

\begin{figure}
    \centering
    \includegraphics[width=3in]{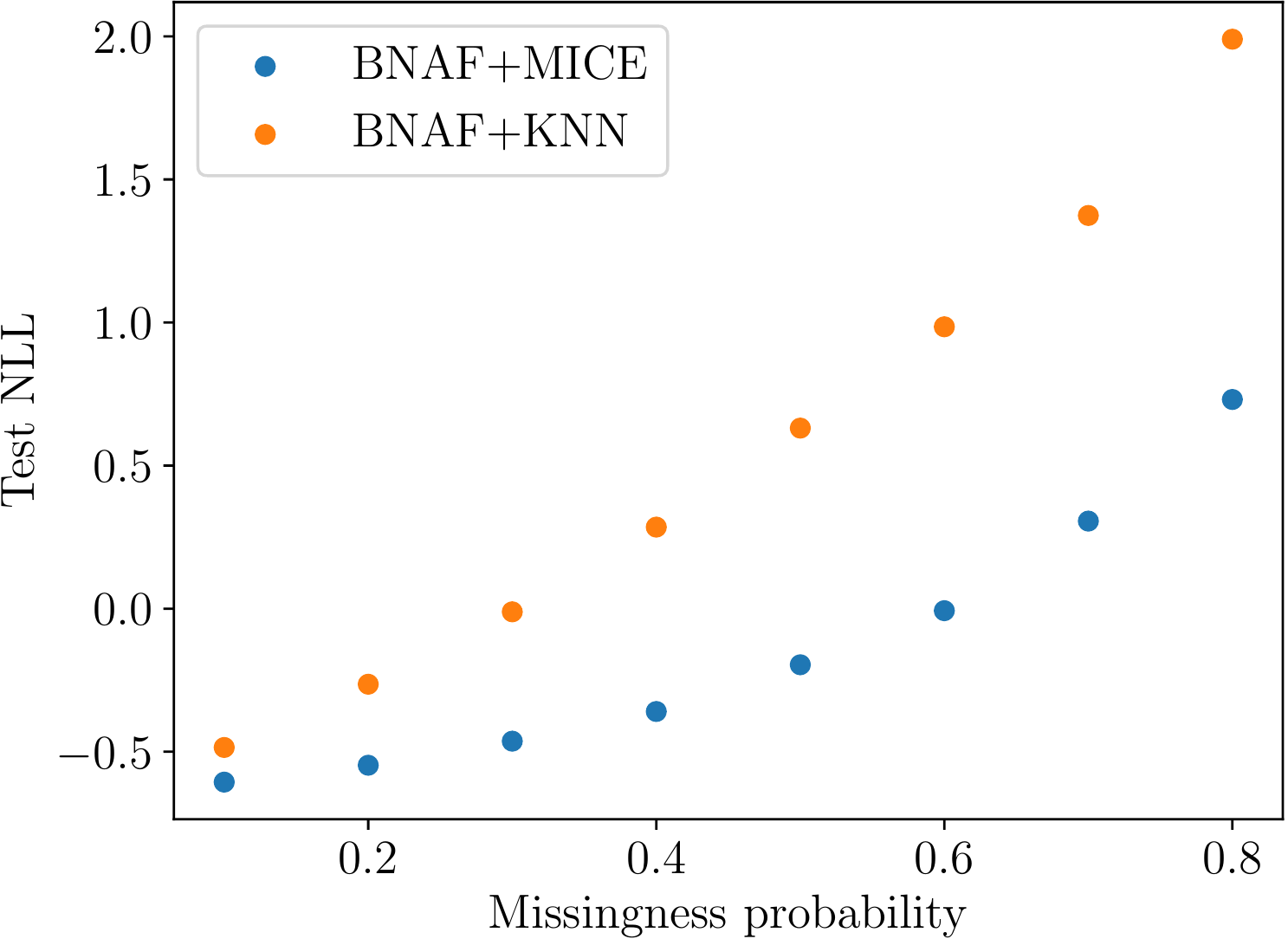}
    \caption{A comparison of data imputation methods on the UCI POWER dataset followed by density estimation with BNAF, showing that MICE imputation outperforms $k$-NN. We subsequently use MICE in the comparison with MDMA in the main text.}
    \label{fig:mice_vs_knn}
\end{figure}

We compare MICE imputation \cite{buuren2010mice} to $k$-NN imputation (with $k=3$ neighbours) \cite{troyanskaya2001missing} on the UCI POWER dataset in \Cref{fig:mice_vs_knn}, before performing density estimation with BNAF \cite{de2020block}. Due to the size of the dataset, we were not able to use $k$-NN imputation on the full 
dataset, but instead split it up into $100$ batches and performed the imputation per batch. Similar results were obtained on the UCI GAS dataset, and for this reason we only compare MDMA to MICE imputation in the main text.

\subsection{Causal discovery}

\begin{figure}
    \centering
    \includegraphics[width=\textwidth]{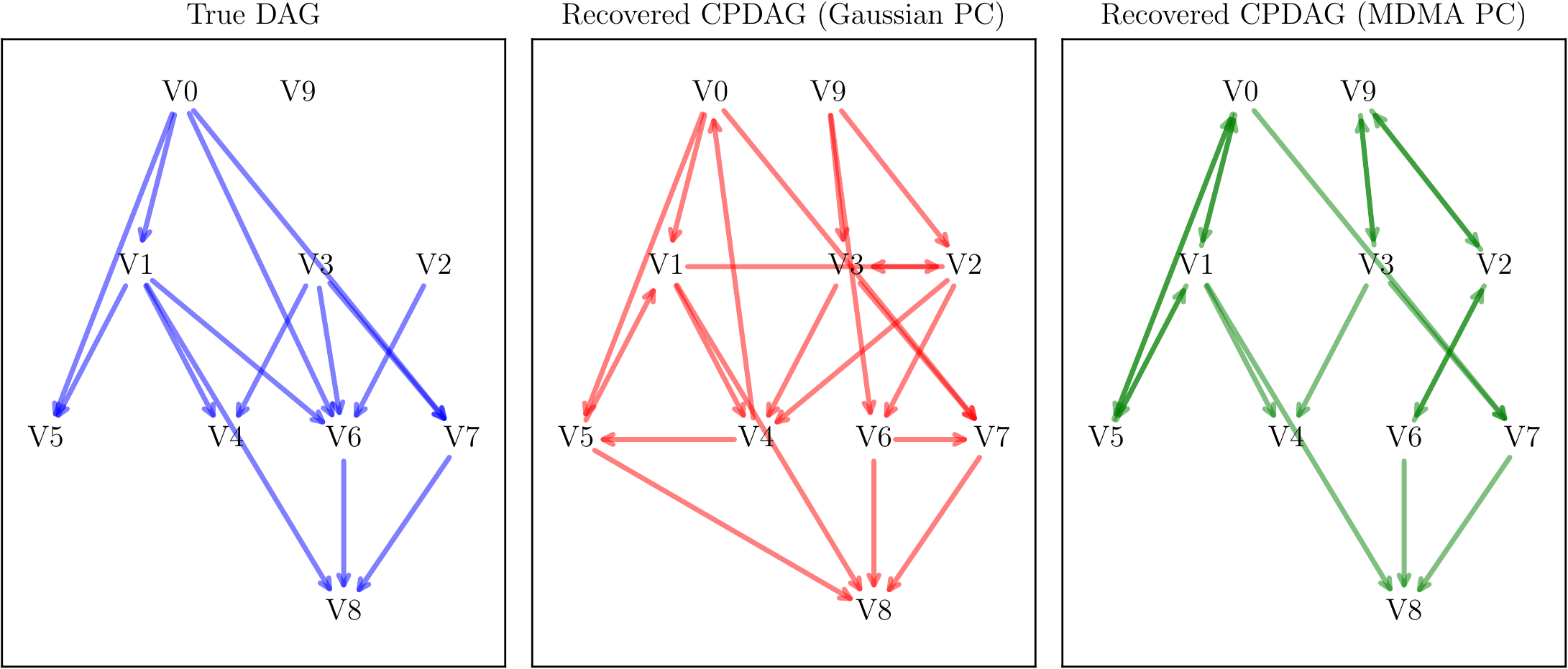}
    \includegraphics[width=\textwidth]{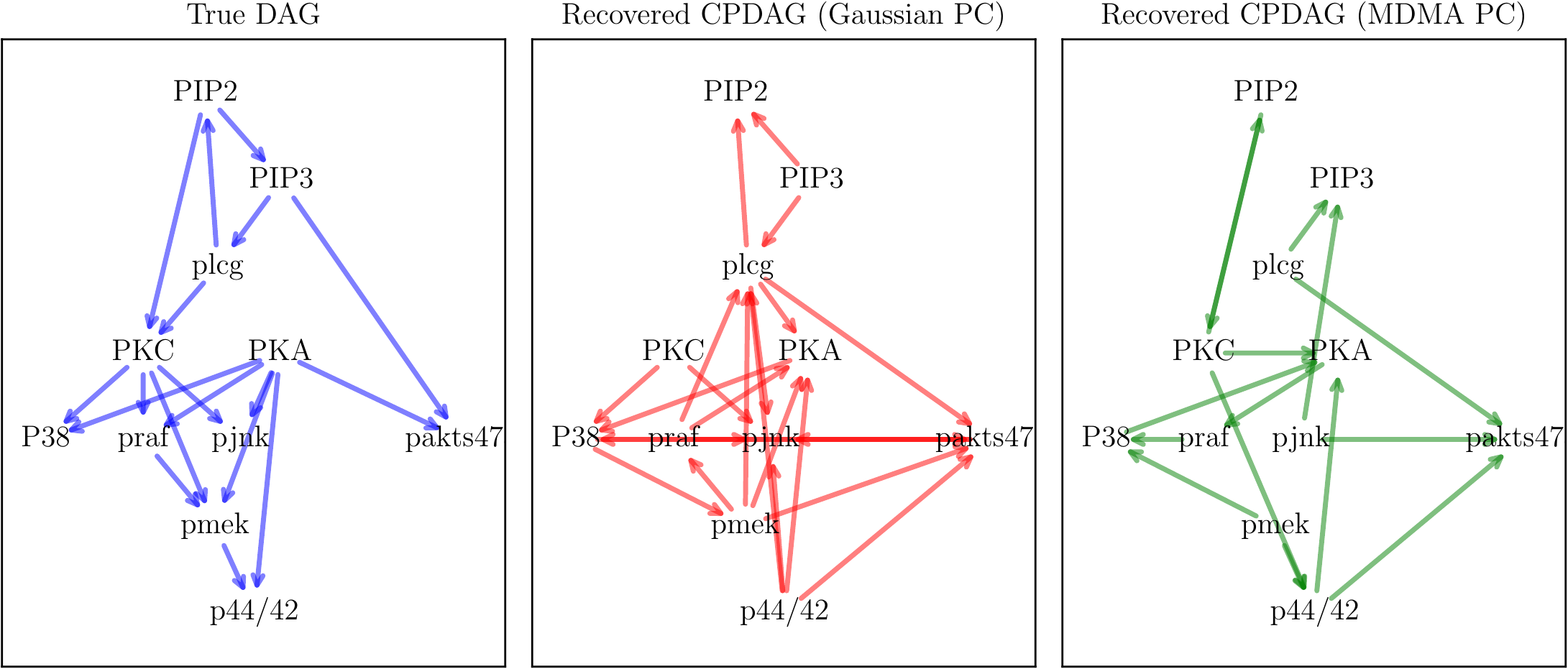}
    \caption{\textbf{Recovered causal graphs}: \textit{Top:} Synthetic data from a random DAG with sigmoidal causality mechanism. The graph inferred using MDMA PC had directional SHD of 11, compared to 15 for the Gaussian PC. \textit{Bottom:} Protein signaling graph \cite{sachs2005causal}. The graph inferred using MDMA PC had directional SHD of 27, compared to 32 for the Gaussian PC.}
    \label{fig:cd_graphs}
\end{figure}

In \Cref{fig:cd_graphs}, we present examples of completely partially directed acyclical graphs (CPDAGs) learned using the PC algorithm, using either MDMA or the vanilla (Gaussian) method for testing conditional independence used in \cite{petersen2021testing}. See \Cref{app:exp_details_CD} for additional details. 

\subsection{Density estimation on real data}

\begin{figure}
    \centering
        \begin{subfigure}[c]{0.45\textwidth}
      \includegraphics[width=\textwidth]{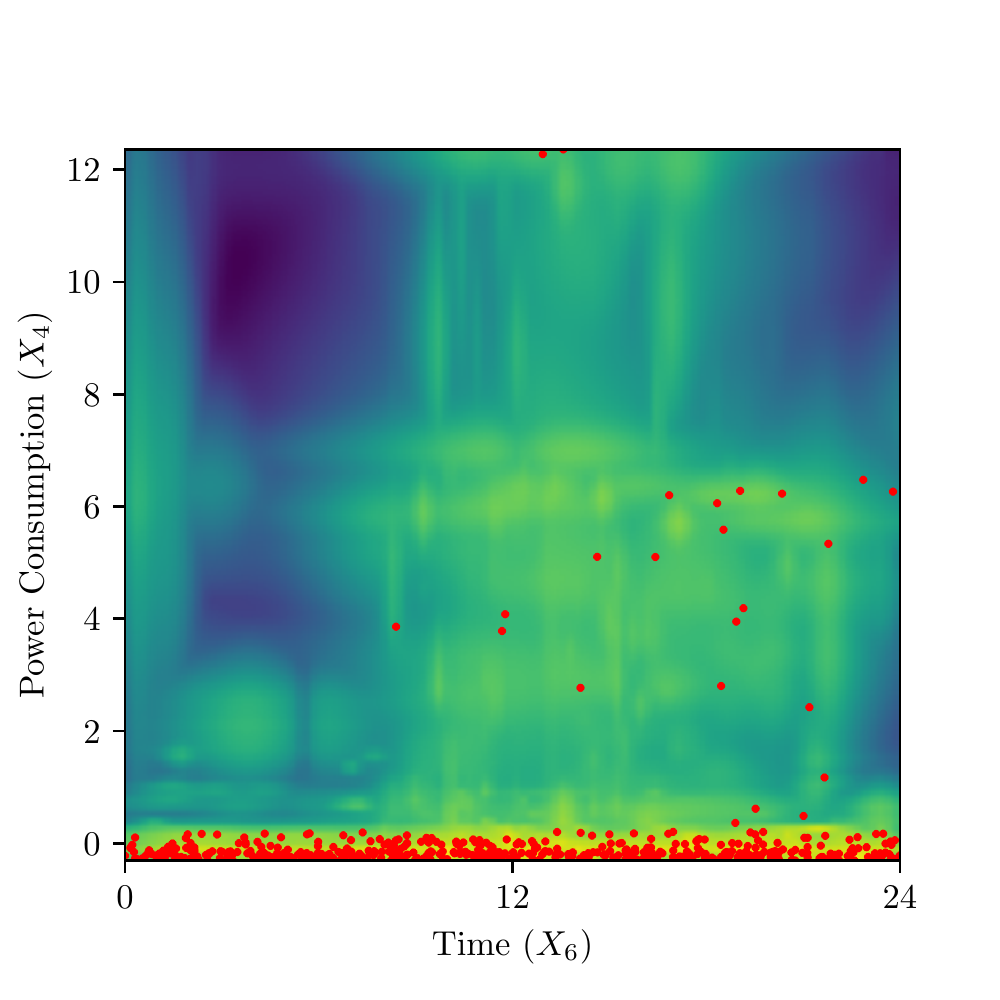}
      \label{}
    \end{subfigure}
    \begin{subfigure}[c]{0.45\textwidth}
    \vspace{0.13in}
      \includegraphics[width=0.91\textwidth]{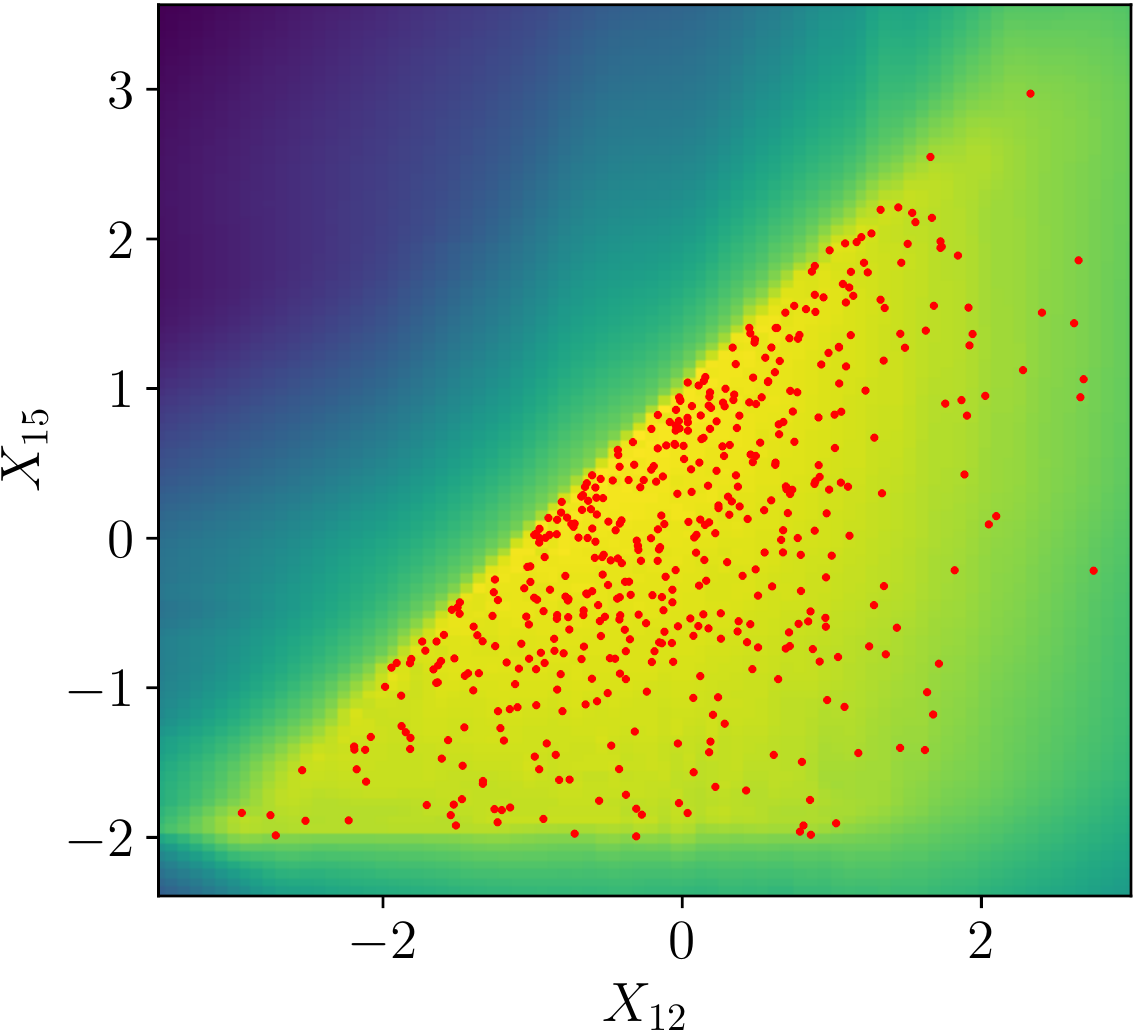}
      \label{}
    \end{subfigure}
    \caption{\textbf{Log marginal density on UCI datasets.} The scatter plot is composed of $500$ samples from the dataset. \textit{Left:} The POWER dataset. One variable corresponds to the time of day, and the other to power consuption from the kitchen of a house. Note the small value of the density during night-time. The data is normalized during training, yet the labels on the horizontal axis reflect the value of the unnormalized variable for interpretability. \textit{Right:} The HEPMASS dataset. Despite MDMA not achieving state-of-the-art results on test likelihood for this dataset, the model still captures accurately the non-trivial dependencies between the variables. }
    \label{fig:power_marginal}
\end{figure}

To demonstrate how MDMA allows one to visualize marginal densities
we show in \Cref{fig:power_marginal} learned bivariate marginals from the 
UCI POWER and HEPMASS datasets. The former is composed of power consumption measurements from different parts of a house, with one of the variables ($X_6$) being the time of day. 

\begin{figure}
    \centering
    \includegraphics[width=3in]{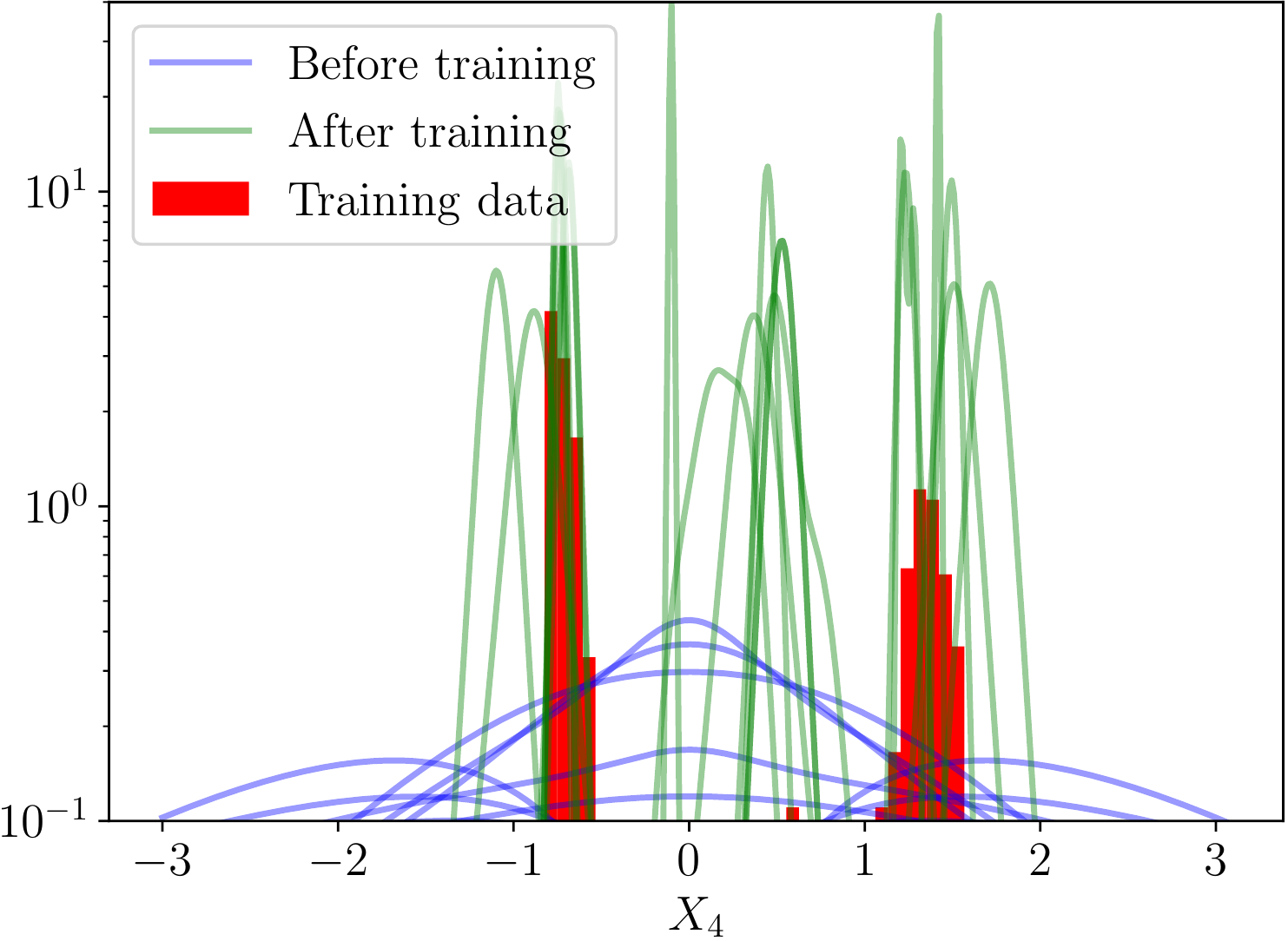}
    \caption{\textbf{Feature learning in MDMA.} We plot ten univariate PDFs $\dot{\varphi}_{ij}$ parameterized as in \cref{app:univariate_marginal_param} for $j=4$, both at initialization and after training on the UCI POWER dataset. Overlaid is a histogram of computed from 500 datapoints. We find that the features localize near the datapoints.}
    \label{fig:phidots}
\end{figure}

\section{Experimental details} \label{app:exp_details}

All experiments were performed on Amazon Web Services using Tesla V-100 GPUs. The total compute time was 3,623 hours, with the vast majority devoted to the experiments on density estimation with missing values (\Cref{sec:demv}), where some of the runs of BNAF required over 72 hours to complete. 

\subsection{Mutual information estimation}
$10^6$ samples from the true density are used for fitting MDMA and for estimating the integral over the log marginals in order to compute the mutual information. The MDMA model used had parameters $r=4, l=5, m=1000$ and was trained with a batch size $500$ and learning rate $0.01$ for $2$ epochs. 

\subsection{Causal discovery} \label{app:exp_details_CD}

In many application areas, causal relationships between random variables can be represented by a directed acyclical graph (DAG).
The PC algorithm \cite{Spirtes2000} is an efficient algorithm for recovering sparse DAGs from observations.
In general, this recovery is complicated by the fact that two DAGs can induce the same probability distribution, leading to them being called Markov equivalent.
Hence, observational data can only help infer the Markov equivalence class of a given DAG.
The equivalence class, known as a completely partially directed acyclical graph (CPDAG, also called essential graph)~\cite{chickering2002learning}, encodes all the dependence information in the induced distribution.
The object of the PC algorithm is therefore the recovery of a CPDAG that is consistent with the data.
This is generally a hard problem, since the cardinality of the space of DAGs is super-exponential in the number of variables \cite{robinson1977counting}. 

The PC algorithm requires repeatedly testing for independence between pairs of variables conditioned on subsets of the remaining variables.
As mentioned in the main text, testing for conditional independence can be reduced to an independence test between variables that depend on conditional 
CDFs~\cite{petersen2021testing}, which can be obtained easily after fitting the joint density using MDMA.
In our experiments, the results of using MDMA as part of the PC algorithm for testing conditional independence are compared to the results obtained by using a Gaussian conditional independence test based on partial correlations. 

The synthetic DAGs were generated using the the Causal Discovery Toolbox.\footnote{https://fentechsolutions.github.io/CausalDiscoveryToolbox/html/index.html} When the sigmoidal causal mechanism is used, given a variable $Y$ and parents $\{X_1,\dots,X_s\}$, then $Y=\underset{i=1}{\overset{r}{\sum}}w_{i}\sigma(X_{i}+w_{0})+\varepsilon$, and if a polynomial mechanism is used then $Y=\varepsilon\left(w_{0}+\underset{i=1}{\overset{s}{\sum}}w_{1i}X_{i}+\underset{i=1}{\overset{s}{\sum}}w_{2i}X_{i}^{2}\right)$, where $w_i,w_{ij}, \varepsilon$ are random. MDMA was trained with $m=1000, L=2, r=3$ for $50$ epochs and learning rate $0.1$ on all datasets. In all experiments we find that the graphs recovered using MDMA are closer to the truth than those recovered using Gaussian PC, as measured by the structural Hamming distance. Example recovered graphs are shown in \Cref{fig:cd_graphs}. 

\subsection{Density estimation on real data}

\begin{table}[]
\caption{Dimension and size of the UCI datasets, and the hyperparameters used for fitting MDMA on these datasets. $m$ is the width of the MDMA model, $l$ and $r$ are respectively the depth and width of the univariate CDF models described in \Cref{app:univariate_marginal_param}. \label{tab:UCI_details} }
\centering 
\begin{tabular}{ccccc}
\toprule
 & POWER   & GAS & HEPMASS & MINIBOONE\\ \hline
d                                               & 6       & 8   & 21      & 43        \\
Training set                                    & 1659917 & 852174                  & 315123                      & 29556                         \\
Validation set                                  & 184435  & 94685                   & 35013                       & 3284                          \\
Test set                                        & 204928  & 105206                  & 174987                      & 3648   \\    \hline
$m$ & 1000 & 4000 & 1000 & 1000 \\ 
$l$ & 2 & 4 & 2 & 2 
\\ 
$r$ & 3 & 5 & 3 & 3
\\
\bottomrule
\end{tabular}
\end{table}

We train MDMA on four UCI datasets, details of the dataset sizes and hyperparameter choices are presented in \Cref{tab:UCI_details}. In all experiments a batch size of $500$ and learning rate of $0.01$ were used. We use the same pre-processing as \cite{papamakarios2017masked}, which involves normalizing and adding noise. Details are provided in the attached code.\footnote{The raw datasets are available for download at https://zenodo.org/record/1161203\#.YLUMImZKjuU} The POWER dataset consists of measurements of power consumption from different parts of a house as well as the time of day. The GAS dataset contains measurements of chemical sensors used to discriminate between different gases. The HEPMASS and MINIBOONE datasets are both measurements from high-energy physics experiments, aiming respectively for the discovery of novel particles and to distinguish between different types of fundamental particles (electron and muon neutrinos).

\section{Tensor decompositions} \label{app:tensor_decomps}

In constructing the MDMA estimator, we are confronted with the problem of combining products of univariate CDFs linearly, in a manner that is both computationally efficient and expressive. 
The linearity constraint reduces this to a tensor decomposition problem (with additional non-negativity and normalization constraints). There is an extensive literature on such efficient tensor decompositions (see \cite{Cichocki2017-gp} for a review).

\begin{figure}
    \centering
     \begin{tikzpicture}
    \node at (-7, 2) (examples){ \includegraphics[scale=0.6] {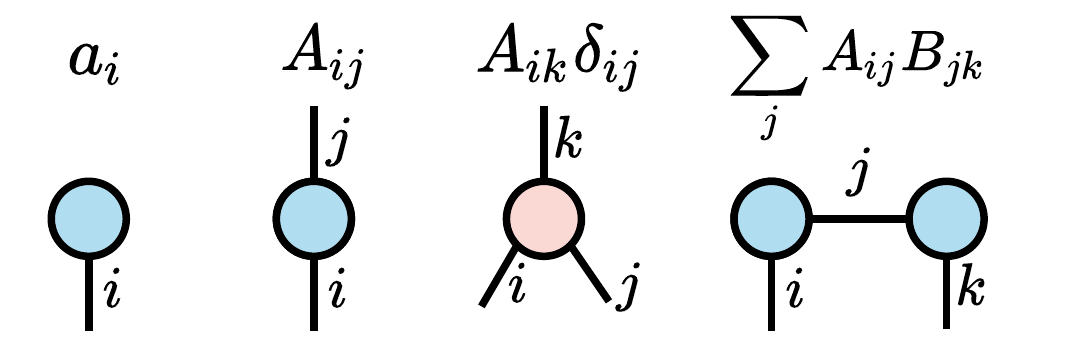}};
    \node[left = 0cm of examples]{a)};
    \node at (-8.5, 0) (phi){ \includegraphics[scale=0.6] {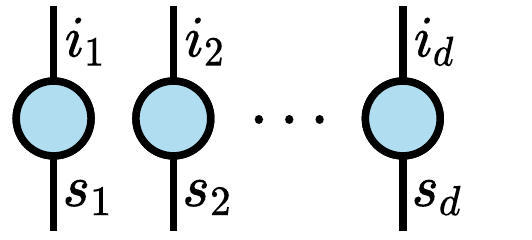}};
    \node[left = 0cm of phi]{b)};
    \node at (-5, 0) (general){ \includegraphics[scale=0.6] {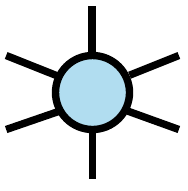}};
    \node[left = 0cm of general]{c)};
        \node at (-1, 1) (HT){ \includegraphics[scale=0.6] {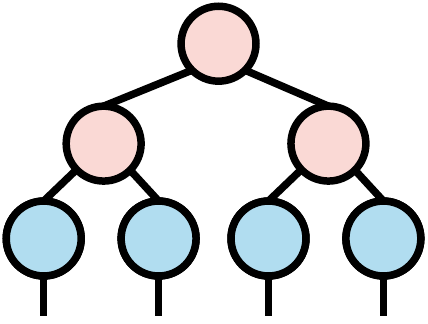}};
    \node[left = 0cm of HT]{d)};
    \end{tikzpicture}
    \caption{\textbf{Tensor decompositions}. a) Tensors of various orders (resp. vectors, matrices, delta tensors). Each edge represents an index, and connecting two edges represents contraction (summation over an index). b) The set of univariate CDFs $\Phi$, which can be viewed as an order $2d$ tensor. c) A general unstructured tensor of order 6. 
    d) The hierarchical Tucker (HT) decomposition \cref{eq:HT}. After suitable normalization, the tensor in d) can be contracted with the tensor $\Phi$ shown in b) to give a multivariate CDF. }
    \label{fig:tensor_decomps}
\end{figure}

The analogy with tensor decompositions becomes clear when we consider discrete rather than continuous variables.
Assume  we wish to model the joint distribution of $d$ discrete variables, each taking one of $S$ possible values.
The distribution is then a function $\varphi_{S}:[S] \times \cdots \times [S] \rightarrow\mathbb{R}$, which can also be viewed as an order $d$ tensor.
A general tensor $\varphi_{S}$ 
will require order of $S^d$ numbers to represent, and is thus impractical even for moderate $d$.
The continuous analog of such a tensor is a multivariate function $F:\mathbb{R}^{d}\rightarrow\mathbb{R}$, with the value of $x_j$ corresponding to the discrete index $s_j$.
We will thus use the same notation for the continuous case. 
A graphical representation of tensors and of the diagonal HT tensor used in MDMA is presented in \Cref{fig:tensor_decomps}. 

\input{arxiv/design_details}
 
\end{document}

%% file: arxiv/header.tex
\usepackage[utf8]{inputenc} 
\usepackage[T1]{fontenc}    
\usepackage{hyperref}       
 \hypersetup{
     colorlinks=true,
     linkcolor=blue,
     filecolor=blue,
     citecolor = red,      
     urlcolor=cyan,
     }
\usepackage{url}            
\usepackage{booktabs}       
\usepackage{amsfonts}       
\usepackage{nicefrac}       
\usepackage{microtype}      
\usepackage{sidecap}

\sidecaptionvpos{figure}{t}

\usepackage{mathtools} 

\usepackage{booktabs} 
\usepackage{tikz} 
\usetikzlibrary{calc,positioning} 
\usepackage{amsfonts}
\usepackage{cleveref}
\usepackage{bm}
\newcommand\subfigwidth{.24\textwidth}
\usepackage{tabu}
\usepackage[ruled,vlined]{algorithm2e}
\SetKw{KwBy}{by}
\usepackage{caption}
\usepackage{subcaption}

\crefname{equation}{}{}

\usepackage{pifont}

\usepackage[compact]{titlesec}

\newcommand{\eq}{\begin{equation*}}
\newcommand{\en}{\end{equation*}}
\newcommand{\eqa}{\begin{eqnarray*}}
\newcommand{\ena}{\end{eqnarray*}}
\newcommand{\eqn}{\begin{equation}}
\newcommand{\enn}{\end{equation}}
\newcommand{\be}{\begin{equation}}
\newcommand{\ee}{\end{equation}}
\newcommand{\eqan}{\begin{eqnarray}}
\newcommand{\enan}{\end{eqnarray}}

\newcommand{\R}{{\mathbb R}}
\newcommand{\N}{{\mathbb N}}

\newcommand{\Fcal}{{\mathcal F}}

\newcommand{\Acal}{{\mathcal A}}

\newcommand{\varphid}{{\dot \varphi}}

\usepackage{dsfont}
\newcommand{\Ind}{\mathds{1}}

\newcommand\independent{\protect\mathpalette{\protect\independenT}{\perp}}
\def\independenT#1#2{\mathrel{\rlap{$#1#2$}\mkern2mu{#1#2}}}

\usepackage{amsthm}

\newtheorem{proposition}{Proposition}

\newtheorem{definition}{Definition}

\newtheorem{remark}{Remark}

%% file: arxiv/intro.tex

\section{Introduction}\label{sec:intro}

Estimating the joint probability density of a set of random variables is a fundamental task in statistics and machine learning that has witnessed much progress in recent years.
While traditional estimators such as histograms~\cite{scott1979optimal,lugosi1996consistency} and kernel-based methods~\cite{Rosenblatt,parzen1962estimation} have appealing theoretical properties and typically perform well in low dimensions, they become computationally impractical above 5-10 dimensions.
Conversely, recent density models based on neural networks~\cite{kingma2018,huang2018neural,oliva2018transformation,grathwohl2018ffjord,de2020block,bigdeli2020learning} scale efficiently with the number of random variables, but lack a crucial feature available to traditional methods: the ability to compute probabilities, marginalize over subsets of variables, and evaluate conditionals. These tasks require integrals of the estimated density, 
which are intractable for modern neural density models.  
Thus, while such operations are central 
to many applications (e.g., inference with missing data, testing for conditional (in)dependence, or performing do-calculus~\cite{pearl_book}), approaches based on neural networks require estimating separate models 
whenever marginals or conditionals are needed.  

Alternatively, one could model the cumulative distribution function (CDF), making computing probabilities and marginalization straightforward. But evaluating the density requires taking $d$ derivatives of the CDF, which incurs an exponential cost in $d$ for a generic computational graph.
This observation has made direct CDF modeling traditionally challenging~\cite{chilinski2020neural}.

In this work, we present the \emph{marginalizable density model approximator} (MDMA), a novel deep network architecture preserving most of the expressive power of neural models for density estimation, while  providing closed form expressions for the probabilities, marginals and conditionals of any subset of the variables.
In a nutshell, the MDMA learns many deep 
scalar representations for each individual variable 
and combines them using hierarchical tensor decompositions \cite{hackbusch2012tensor,cichocki2016tensor} into a tractable multivariate CDF that can be fitted using stochastic gradient descent.
Additionally, sampling from MDMA can be parallelized along the input dimension, resulting in 
a very low space complexity and a time complexity that scales only logarithmically with the number of variables in the problem (as opposed to linearly in naive autoregressive sampling, see below in \Cref{sec:related}). 

As could be expected, the architectural choices that allow for easy marginalization take  a minor toll  in terms of performance. Indeed, while competitive, our model admittedly does not beat  state-of-the-art models in out-of-sample log-likelihood of 
high-dimensional datasets. On the other hand, it does 
beat those same models in a task for which the latter are ill-prepared: learning densities from data containing missing values, a
common setting in some application areas such as genomics~\cite{li2009genotype, marchini2010genotype}. 
While our model is able to deal optimally with missing values by evaluating, for every data point,  the marginal likelihood over its non-missing values, other models must resort to data
imputation. Consequently, we significantly outperform state-of-the-art neural density estimators trained 
using a number of common data imputation strategies. We also show MDMA can be used to test for conditional independence, which is useful for discovering the causal structure in graphs, a task on which it outperforms existing methods, and that it enables estimation of mutual information between arbitrary subsets of variables after fitting a single model. Additionally, we prove that the model class is a universal approximator over the space of absolutely continuous multivariate distributions.

The structure of this paper is as follows. In \Cref{sec:related} we 
review related works. In \Cref{sec:marginalizable_models} we
present our new model and its theoretical properties.
We present our experimental results in~\Cref{sec:experiments}, and 
conclude in~\Cref{sec:discussion}.

%% file: arxiv/litterature.tex

\section{Related work}
\label{sec:related}

Modern approaches to non-parametric density estimation, 
based on normalizing flows~\cite{kingma2018,huang2018neural,oliva2018transformation,grathwohl2018ffjord,de2020block,bigdeli2020learning} (see~\cite{kobyzev2020normalizing,papamakarios2021normalizing} for recent reviews), model expressive yet invertible functions that transforms a simple (usually uniform or Gaussian) density to the target density.
Nonetheless, as previously mentioned, such architectures 
lead to intractable derived quantities such as probabilities, marginals and/or conditional distributions.

Moreover, many normalizing flow models rely on an autoregressive construction, 
which makes the cost of generating samples scale linearly with the dimension.
This can be circumvented using inverse autoregressive flows~\cite{kingma2016improving}, but in this dual case a linear cost 
is incurred instead in density evaluations and hence in training. Another solution is training a feed-forward network using the outputs of a trained autoregressive model~\cite{oord2018parallel}. With MDMA, fast inference and sampling is achieved without requiring this distillation procedure.

Tensor decompositions~\cite{hackbusch2012tensor,cichocki2016tensor}, 
which are exploited in this work, have been used in various applications 
of signal processing, machine learning, computer vision, and more~\cite{vasilescu2002multilinear,cichocki2009nonnegative,anandkumar2014tensor,papalexakis2016tensors,sidiropoulos2017tensor}.
Recently, such decompositions have been used to speed-up or reduce the number of parameters in existing deep architectures~\cite{lebedev2014speeding,tai2015convolutional,novikov2015tensorizing,kim2015compression,yang2016deep,chen2018sharing}.
In addition to their practical appeal, 
tensor methods have been widely studied to understand the success of deep neural networks \cite{cohen2016expressive,haeffele2015global,janzamin2015generalization,janzamin2015beating,sharir2017expressive}


%% file: arxiv/mdma2.tex

\section{Marginalizable Density Models}
\label{sec:marginalizable_models}

\subsection{Notations}
In the following, we use a capital and lowercase Roman letter (e.g., $F$ and $f$) or Greek letters along with dot above (e.g., $\varphi$ and $\varphid$) to denote respectively absolutely continuous CDFs of arbitrary dimensions and the corresponding densities. When dealing with multivariate distributions, the marginal or conditional distribution over a subset of variables will be indicated by the argument names (i.e., $F(x_{1}|x_{2},x_{3})=\mathbb{P}\left[X_{1}\leq x_{1}|X_{2}=x_{2},X_{3}=x_{3}\right]$).

For a positive integer $p \in \N \setminus 0$, 
let $[p] = \{1, \dots, p\}$.
Denote the space of absolutely continuous univariate and $d$-dimensional CDFs respectively by $\Fcal_1$ and $\Fcal_d$. 
For any $F \in \Fcal_1$, the density $f:\R \to \R_{+}$ is $f(x) = \partial F(x)/\partial x$.
Similarly, for any $F \in \Fcal_d$, the density $f(\bm x):\R^d \to \R_{+}$ is $f(\bm x) = \partial^d F(\bm x)/\partial x_1 \cdots \partial x_d$,
and $F_j(x) = \lim_{z \to \infty} F(z, \dots, x, \dots, z) \in \Fcal_1$ for $j \in [d]$ is the $j$th marginal distribution.

\subsection{The bivariate case}
For the task of modeling joint distributions of two variables supported on $\mathbb{R}^{2}$,
consider a family of univariate CDFs
$\{\varphi_{i,j}\,\}_{i\in[m],\,j\in[2]}$ with $\varphi_{i,j} \in \Fcal_1$, i.e., the functions 
$\varphi_{i,j}:\mathbb{R} \rightarrow [0,1]$ satisfy
\begin{align*}
\lim_{x\to -\infty}\varphi_{i,j}(x) = 0 \,, \qquad     
\lim_{x\to \infty}\varphi_{i,j}(x) = 1 \,, \qquad     
\varphid_{i,j}(x) = \partial \varphi(x) / \partial x  \geq 0 \,.
\end{align*}
These functions are our basic building block, 
and we model them using a simple neural architecture proposed in~\cite{balle2018variational} and described in~\Cref{sec:theory}. 
If $A$ is an $m \times m$ matrix of nonnegative elements satisfying $\sum_{i,j=1}^m A_{i,j} =1$,
we can combine it with the univariate CDFs to obtain
\begin{align} 
\label{eq:bivariate_cdf}
    F(x_{1},x_{2})=\sum\limits_{i,j=1}^{m} A_{i,j}\varphi_{i,1}(x_{1})\varphi_{j,2}(x_{2}).
\end{align}
The coefficients $A_{i,j}$ encode the dependencies between the two variables, and the normalization ensures that $F$ is a valid CDF, that is $F \in \Fcal_2$.
Even though in each summand the interaction is modeled by a single scalar parameter, such a model can be used to approximate well complex interactions between $x_1$ and $x_2$ if $m$ is sufficiently large, as we show in \Cref{sec:theory}. 
The advantage of this construction is that $\{\varphid_{i,j}\}_{i\in[m],\,j\in[2]}$, the family of densities corresponding to the univariate CDFs, leads immediately to 
\begin{align*}
f(x_{1},x_{2})=\sum\limits_{i,j=1}^{m}A_{i,j}\varphid_{i,1}(x_{1})\varphid_{j,2}(x_{2}). 
\end{align*}
It is similarly straightforward to obtain marginal and conditional quantities, e.g.:
\begin{align*}
F(x_{1})&=\sum\limits_{i,j=1}^{m}A_{i,j}\varphi_{i,1}(x_{1}), \qquad
    F(x_{1}|x_{2})=\frac{\sum\limits_{i,j=1}^{m}A_{i,j}\varphi_{i,1}(x_{1})\varphid_{j,2}(x_{2})}{\sum\limits_{i,j=1}^{m}A_{i,j}\varphid_{j,2}(x_{2})}, 
\end{align*}
and the corresponding densities result from replacing $\varphi_{i,1}$ by $\varphid_{i,1}$.
Deriving these simple expressions 
relies on the fact that \eqref{eq:bivariate_cdf} combines the univariate CDFs linearly. Nonetheless, it is clear that, with $m \to \infty$ and for judiciously chosen univariate CDFs, such a model is a universal approximator of both CDFs and sufficiently smooth densities.


\subsection{The multivariate case}\label{sec:multivariate}

To generalize the bivariate case, consider a collection of univariate CDFs $\{\varphi_{i,j}\}_{i\in[m],\,j\in[d]}$ with $\varphi_{i,j} \in \Fcal_1$ for each $i$ and $j$, and define the tensor-valued function $\Phi \colon \R^d \times [m]^d \to [0,1]$ by $\Phi(\bm x)_{i_1, \dots, i_d} = \prod_{j=1}^{d}\varphi_{i_{j},j}(x_{j})$ for $\bm x \in \R^d$.
Furthermore, denote the class of normalized order $d$ tensors with $m$ dimensions in each mode and nonnegative elements by
\begin{align}
    \Acal_{d,m} = \{A \in \R^{m \times \cdots \times m}\colon A_{i_1, \dots, i_d} \ge 0,\, \sum\limits_{i_1, \dots, i_d = 1}^m A_{i_1, \dots, i_d} = 1 \}.
    \label{eq:define_acal}
\end{align}
\begin{definition}[Marginalizable Density Model Approximator]
For $\Phi \colon \R^d \times [m]^d \to [0,1]$ as above and $A \in \Acal_{d,m}$,
the marginalizable density model approximator (MDMA) is
\begin{align} \label{eq:F_general_A}
    F_{A,\Phi}(\bm x)= \langle A, \Phi(\bm x) \rangle  = \sum\limits_{i_{1},\dots,i_{d}=1}^{m}A_{i_{1},\dots,i_{d}}\prod\limits_{j=1}^{d}\varphi_{i_{j},j}(x_{j}).
\end{align}
\end{definition}
If is clear that the conditions on $\Phi$ and $A$ imply that $F_{A,\Phi} \in \Fcal_d$.  
As in the bivariate case, densities or marginalization over $x_j$ are obtained by replacing each $\varphi_{i,j}$ respectively by $\varphid_{i,j}$ or $1$. 
As for conditioning, considering any disjoint subsets $R=\left\{ k_{1},\dots,k_{r}\right\}$ and $S=\left\{ j_{1},\dots,j_{s}\right\}$ of $[d]$ such that $R \cap S = \emptyset $, we have  
\begin{align*}
F_{A, \Phi}(x_{k_{1}},\dots,x_{k_{r}}|x_{j_{1}},\dots,x_{j_{s}}) & = \frac{\sum\limits_{i_{1},\dots,i_{d}=1}^{m}A_{i_{1},\dots,i_{d}}\underset{k\in R}{\prod}\varphi_{i_{k},k}(x_{k})\underset{j\in S}{\prod}\varphid_{i_{j},j}(x_{j})}{\sum\limits_{i_{1},\dots,i_{d}=1}^{m}A_{i_{1},\dots,i_{d}}\underset{j\in S}{\prod}\varphid_{i_{j},j}(x_{j})}.
\end{align*}

For a completely general tensor $A$
with $m^d$ parameters, the expression \eqref{eq:F_general_A} is computationally impractical, hence some structure must be imposed on $A$.
For instance, one simple choice is $A_{i_1, \dots, i_d}=a_{i_1} \delta_{i_1, \dots, i_d}$, which leads to $F_{A,\Phi}(\bm x)=\sum_{i=1}^m a_{i}\prod_{j=1}^{d}\varphi_{i,j}(x_{j})$, with $a_i \ge 0$ and $\sum_{i=1}^m a_{i} = 1$.
Instead of this top-down approach, $A$ can be tractably constructed bottom-up, as we explain next. 

Assuming 
$d=2^p$ for integer~$p$, 
define $\varphi^{\ell}\colon \R^d \to \R^{m \times 2^{p-\ell+1}}$
for $\ell \in \{ 1, \dots, p \}$ recursively by
\begin{align}\label{eq:HT_recurs}
\varphi_{i,j}^{\ell}(\bm{x}) & = 
\begin{cases}
\varphi_{i,j}(x_j), \qquad &\ell = 1 \\
\sum\limits _{k=1}^{m}\lambda_{i,k,j}^{\ell-1}\varphi_{k,2j-1}^{\ell-1}(\bm{x})\varphi_{k,2j}^{\ell-1}(\bm{x}), \qquad &\ell  = 2, \dots, p
\end{cases}
\end{align} 
for $i \in [m]$, $j \in [2^{p-\ell +1}]$, and where $\lambda^{\ell}$ is a non-negative $m \times m \times 2^{p-\ell+1}$ tensor, normalized as $\sum_{k=1}^m \lambda_{i,k,j}^{\ell}=1$. 
The joint CDF can then be written as
\begin{align} 
\label{eq:HT}
 F_{A^{\mathrm{HT}}, \Phi}(\bm x) &= \sum\limits_{k=1}^m \lambda_{k}^{p} \varphi_{k,1}^{p}(\bm x)\varphi_{k,2}^{p}(\bm x),
\end{align}
with $\lambda^p \in \mathbb{R}_{+}^{m}$ satisfying $\sum_{k=1}^m \lambda_k^p=1$. 
It is easy to verify that the underlying $A^{\mathrm{HT}}$ satisfies $A^{\mathrm{HT}} \in \Acal_{d,m}$ defined in \eqref{eq:define_acal}. A graphical representation of this tensor is provided in \Cref{fig:tensor_decomps} in the supplementary materials.  

For example, for $d=4$, we first combine $(x_1,x_2)$ and $(x_3,x_4)$ into
\begin{align*}
    \varphi^2_{i,1}(\bm{x}) = \sum_{k=1}^m \lambda^1_{i,k,1} \varphi_{k,1}(x_1) \varphi_{k,2}(x_2) \,,
    \qquad 
    \varphi^2_{i,2}(\bm{x}) = \sum_{k=1}^m \lambda^1_{i,k,2} \varphi_{k,3}(x_3) \varphi_{k,4}(x_4) \,, 
\end{align*}
and then merge them as 
\begin{align}
F_{A^{\mathrm{HT}},\Phi} (\bm{x}) =    \sum_{k=1}^m \lambda_k^2 \varphi^2_{k,1}(\bm{x}) \varphi^2_{k,2}(\bm{x}),
\end{align}
from which we can read off that $A^{\mathrm{HT}}_{i_{1},i_{2},i_{3},i_{4}}=\sum_{k=1}^{m}\lambda_{k}^2\lambda_{k,i_{1},1}^{1}\lambda_{k,i_{3},2}^{1}\delta_{i_{1},i_{2}}\delta_{i_{3},i_{4}}$. 

Note that the number of parameters required to represent $A^{\mathrm{HT}}$ is only $\mathrm{poly}(m,d)$. The construction is easily 
generalized to dimensions $d$ not divisible by 2. 
Also, the number of $\varphi$ factors combined at 
each iteration in (\ref{eq:HT_recurs}) (called the \emph{pool size}),
 can be any positive integer.
 This construction is a variant of the hierarchical Tucker decomposition of tensors~\cite{Hackbusch2009-yr}, which has been used in the construction of tensor networks for image classification in \cite{cohen2016expressive}.  

Given a set of training points $\{\bm{x} \}_{i=1}^{N}$, we fit
MDMA models by maximizing the log of the density with respect to both the parameters in $\{ \varphi_{i,j}\}_{i\in[m],j\in[d]}$ and the components of $A$. We present additional details regarding the choice of architectures and initialization in \Cref{app:design_details}.

\subsection{A non-marginalizable MDMA} \label{sec:nmdma}

We can construct a more expressive variant of MDMA at the price of losing the ability to marginalize and condition on arbitrary subsets of variables. We find that the resulting model leads to state-of-the-art performance on a density estimation benchmark.
We define $\mathbf{v}=\mathbf{x}+\mathbf{T}\sigma(\mathbf{x})$ where $\mathbf{T}$ is an upper-triangular matrix with non-negative entries and $0$ on the main diagonal. Note that $\left|\frac{\partial\mathbf{v}}{\partial\mathbf{x}}\right|=1$.
Given some density $f(v_{1},\dots,v_{d})=\prod\limits_{j=1}^{d}\varphid_{j}(v_{j})$, we have 
\begin{align}
f(x_{1},\dots,x_{d})=\left|\frac{\partial\mathbf{v}}{\partial\mathbf{x}}\right|f(v_{1},\dots,v_{d})=f(v_{1}(\mathbf{x}),\dots,v_{d}(\mathbf{x})).
\end{align}
We refer to this model nMDMA, since it no longer enables efficient marginalization and conditioning. 

\subsection{MDMA sampling}

Given an MDMA as in \cref{eq:F_general_A}, we can sample in the same manner as for autoregressive models:
from $u_{1},\dots,u_{d}$ independent $U(0,1)$ variables, we obtain a sample from $F_{A,\Phi}$ by computing 
\begin{align*}
x_{1} =F_{A,\Phi}^{-1}(u_{1}),\quad x_{2} =F_{A,\Phi}^{-1}(u_{2}|x_1) \quad \cdots \quad x_{d} = F_{A,\Phi}^{-1}(u_{d}|x_{1},\dots,x_{d-1}),
\end{align*}
where, unlike with autoregressive model, the order of taking the conditionals does not need to be fixed.
The main drawback of this method is that due to the sequential nature of the sampling the computational cost is linear in $d$ and cannot be parallelized.

However, the structure of $F_{A,\Phi}$ can be leveraged to sample far more efficiently.
Define by $R_{A}$ a vector-valued categorical random variable taking values in $\left[m\right] \times \cdots \times \left[m\right]$, with distribution 
\begin{align*}
    \mathbb{P}\left[R_{A}=(i_{1},\dots,i_{d})\right]=A_{i_{1},\dots,i_{d}}.
\end{align*}
The fact that $A \in \Acal_{d,m}$ with $\Acal_{d,m}$ from \eqref{eq:define_acal} ensure the validity of this definition.
Consider a vector $(\tilde{X}_{1},\dots,\tilde{X}_{d},R_{A})$ where $R_A$ is distributed as above, and 
\begin{align*}
    \mathbb{P}\left[\tilde{X}_{1}\leq x_{1},\dots,\tilde{X}_{d}\leq x_{d}|R_{A}=\mathbf{r}\right]=\prod\limits_{i=1}^{d}\varphi_{r_{i},i}(x_{i}),
\end{align*}
for the collection $\{\varphi_{i,j}\}$ of univariate CDFs.
Denoting the distribution of this vector by $\tilde{F}_{A,\Phi}$, marginalizing over $R_A$ gives 
\begin{align*}
	\tilde{F}_{A,\Phi}(\bm x)=\underset{\mathbf{r}\in[m]^{d}}{\sum}\mathbb{P}\left[R_{A}=\mathbf{r}\right]\prod\limits_{j=1}^{d}\varphi_{r_{j},j}(x_{j})
	=F_{A,\Phi}(\bm x).
\end{align*}
Instead of sampling directly from the distribution $F_{A,\Phi}$, we can thus sample from $\tilde{F}_{A,\Phi}$ and discard the sample of $R_A$.
To do this, we first sample from the categorical variable $R_A$. Denoting this sample by $\mathbf{r}$, we can sample from $\tilde{X}_{i}$ by inverting the univariate CDF $\varphi_{r_{i},i}$.
This can be \textit{parallelized} over $i$. 

The approach outlined above is impractical since $R_{A}$ can take $m^d$ possible values, yet if $A$ can be expressed by an efficient tensor representation this exponential dependence can be avoided. 
Consider the HT decomposition \cref{eq:HT}, which can be written as 
\begin{align}\label{eq:HT_mixture}
F_{A^{\mathrm{HT}},\Phi}(\bm{x})=\left\langle A^{\mathrm{HT}},\Phi(\bm{x})\right\rangle =\prod\limits _{\ell=1}^{p}\prod\limits _{j_{\ell}=1}^{2^{p-\ell}}\sum\limits _{k_{\ell,j_{\ell}}=1}^{m}\lambda_{k_{\ell+1,\left\lceil j_{\ell}/2\right\rceil },k_{\ell,j_{\ell}},j_{\ell}}^{\ell}\Phi(\bm{x})_{k_{1,1},\dots,k_{1,d/2}},
\end{align}
that is a normalized sum of $O(md)$ univariate CDFs. 
\begin{proposition}\label{prop:HT_sampling}
Sampling from \eqref{eq:HT_mixture} can be achieved in $O(\log d)$ time requiring the storage of only $O(d)$ integers.
\end{proposition}
Note that the time complexity of this sampling procedure depends only logarithmically on $d$.
The reason is that a simple hierarchical structure of Algorithm \ref{algo:ht_sampling}, where $\textrm{Mult}$ denotes the multinomial distribution.
\begin{figure}[ht]
\vspace{-.2in}
  \centering
  \begin{minipage}{.7\linewidth}
  \begin{algorithm}[H]
    \SetAlgoLined
    \KwResult{$x_j$ for $j= 1, \dots, d$}
    $k_{p,1}  \sim \textrm{Mult}(\lambda_{*}^p)$\;
     \For{$\ell \gets p - 1$ \KwTo $1$ \KwBy $-1$}{
      $ k_{\ell,j} \sim \textrm{Mult}(\lambda^{\ell}_{k_{\ell + 1, \left\lceil j/2\right\rceil }, *,j}), \qquad j = 1, \dots, 2^{p - \ell}$\;
     }
     $ x_j \sim \varphi_{k_{1, \left\lceil j/2\right\rceil},j}, \qquad j=1, \dots, 2^{p}$\;
     \caption{Sampling from the HT MDMA}\label{algo:ht_sampling}
    \end{algorithm}
    \end{minipage}
    \vspace{-.2in}
    \label{alg:HT}
\end{figure}

The logarithmic dependence is only in the sampling from the categorical variables, which is inexpensive to begin with.
We thus avoid the linear dependence of the time complexity on $d$ that is common in sampling from autoregressive models. Furthermore, the additional memory required for sampling scales like $\log m$ (since storing the categorical samples requires representing integers up to size $m$), and aside from this each sample requires evaluating a single product of univariate CDFs (which is independent of $m$). In preliminary experiments, we have found that even for densities with $d \leq 10$, this sampling scheme is faster by $1.5$ to $2$ orders of magnitude than autoregressive sampling. The relative speedup should only increase with $d$.

%% file: arxiv/theory.tex

\subsection{Universality of the MDMA}
\label{sec:theory}

To model functions in $\Fcal_1$, we use $\Phi_{l,r,\sigma}$, a class of constrained feedforward neural networks proposed in \cite{balle2018variational} with $l$ hidden layers, each with $r$ neurons, and $\sigma$ a nonaffine, increasing and continuously differentiable elementwise activation function, defined as
\begin{align*}
    \Phi_{l,r,\sigma} = \{ \varphi \colon \R \to [0,1],\, \varphi(x) = \mbox{sigmoid} \circ L_{l} \circ \sigma \circ L_{l-1} \circ \sigma \cdots \circ \sigma \circ L_1 \circ \sigma \circ L_0(x)\},
\end{align*}
where $L_i \colon \R^{n_{i}} \to \R^{n_{i+l}}$ is the affine map $L_i(x) = W_i x + b_i$ for an $n_{i+1} \times n_{i}$ weight matrix $W_i$ with \emph{nonnegative} elements and an $n_{i+1} \times 1$ bias vector $b_i$, with $n_{l+1} = n_0 = 1$ and $n_i = r$ for $i \in [l]$.
The constraints on the weights and the final sigmoid guarantee that $ \Phi_{l,r,\sigma} \subseteq \Fcal_1$, and for any $\varphi \in \Phi_{l,r,\sigma}$, the corresponding density $\varphid (x) = \partial \varphi(x) / \partial x$ can be obtained with the chain rule.
The universal approximation property of the class $\Phi_{l,r,\sigma}$ is expressed in the following proposition.
\begin{proposition}\label{prop:universal_1}
$\cup_{l,r} \Phi_{l,r,\sigma}$ is dense in $\Fcal_1$ with respect to the uniform norm.
\end{proposition}
While the proof in the supplementary assumes that $\lim_{x\to-\infty}\sigma(x) = 0$ and $\lim_{x\to\infty} \sigma(x) = 1$, it can be easily modified to cover other activations.
For instance, in our experiments, we use $\sigma(x) = x + a \tanh(x)$ following \cite{balle2018variational}, and refer to the supplementary material for more details regarding this case.
In the multivariate case, consider the class of order $d$ tensored-valued functions with $m$ dimensions per mode defined as
\begin{align*}
    \Phi_{m,d,l,r,\sigma} = \{ \Phi \colon \R^d \times [m]^d \to [0,1], \, \Phi(\bm x)_{i_1, \dots, i_d} = \textstyle \prod\limits_{j=1}^{d}\varphi_{i_{j},j}(x_{j}),\, \varphi_{i,j} \in \Phi_{l,r,\sigma}\}.
\end{align*}
Combining $\Phi_{m,d,l,r,\sigma}$ with the $\Acal_{d,m}$, the normalized tensors introduced in \Cref{sec:multivariate}, the class of neural network-based MDMAs can then be expressed as
\begin{align*}
    \textrm{MDMA}_{m,d,l,r,\sigma} = \{ F_{A,\Phi}\colon\R^d \to [0,1],\,F_{A,\Phi}(\bm x)= \langle A, \Phi(\bm x) \rangle, \, A \in \Acal_{d,m},\, \Phi \in  \Phi_{m,d,l,r,\sigma} \}.
\end{align*}
\begin{proposition}\label{prop:universal_d}
The set $\cup_{m,l,r} \textrm{MDMA}_{m,d,l,r,\sigma}$ is dense in $\Fcal_{d}$ with respect to the uniform norm.
\end{proposition}
The proof relies on the fact that setting $m=1$ yields a class that is dense in the space of $d$-dimensional CDFs with independent components. All proofs are provided in \Cref{app:proofs}. 

%% file: arxiv/exp_toy_combined.tex

\subsection{Toy density estimation}
\begin{figure}[h]
  \centering
  \renewcommand{\tabcolsep}{1pt}
  \begin{tikzpicture}
    \node {\begin{tabular}[c]{cccc}
\begin{subfigure}[c]{\subfigwidth}
      \includegraphics[width=\textwidth]{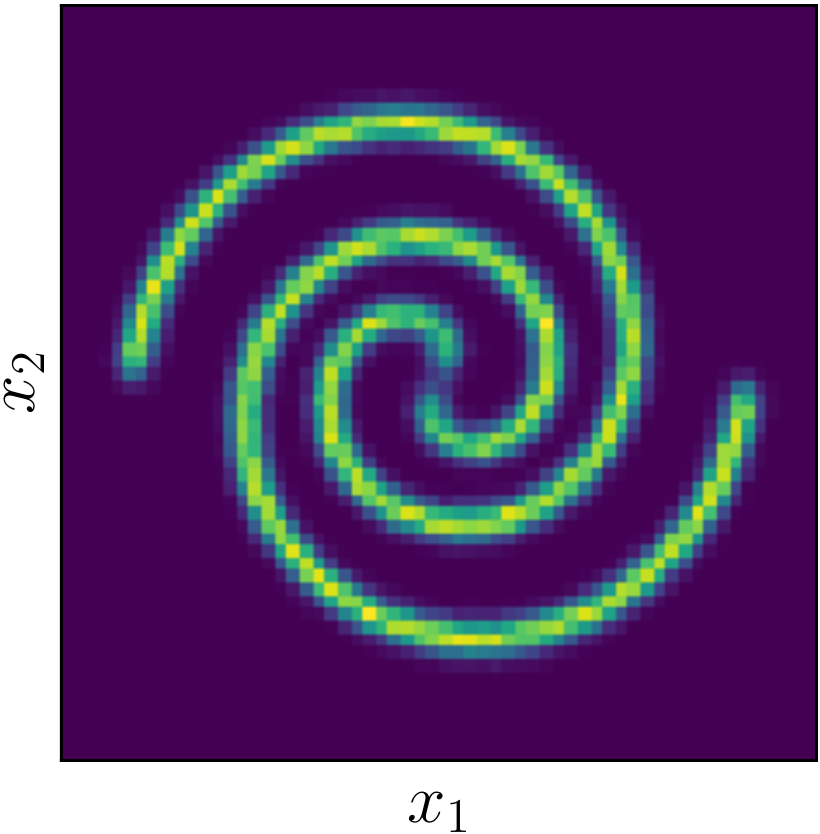}
      \label{}
    \end{subfigure}&
    \begin{subfigure}[c]{\subfigwidth}
      \includegraphics[width=\textwidth]{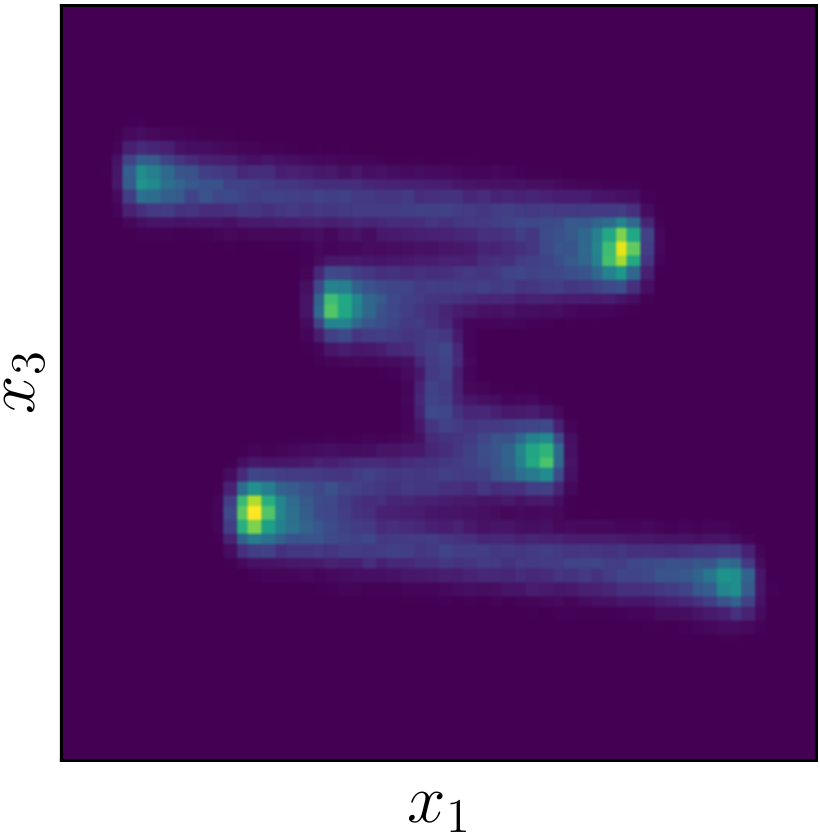}
      \label{}
    \end{subfigure}&
    \begin{subfigure}[c]{\subfigwidth}
      \includegraphics[width=\textwidth]{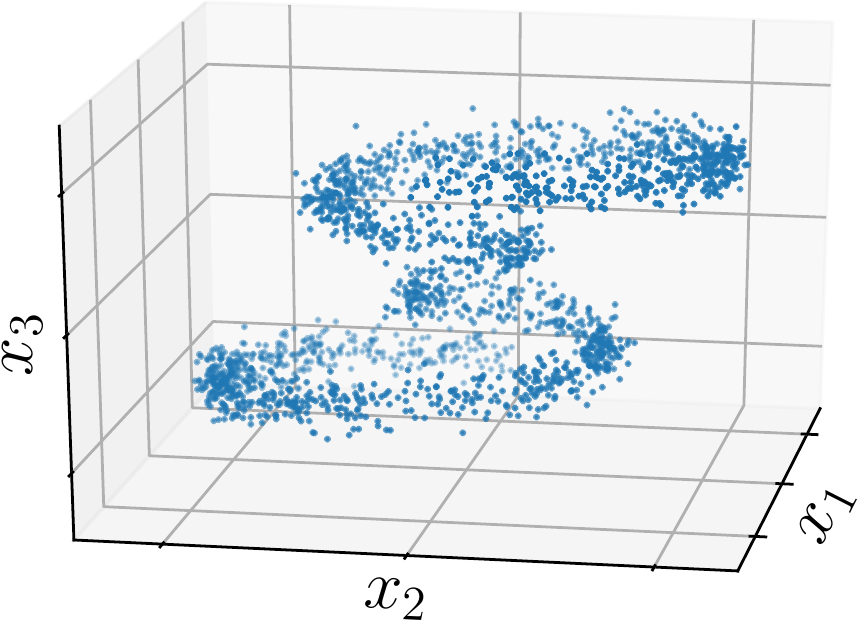}
      \label{}
    \end{subfigure}&
    \begin{subfigure}[c]{\subfigwidth}
      \includegraphics[width=\textwidth]{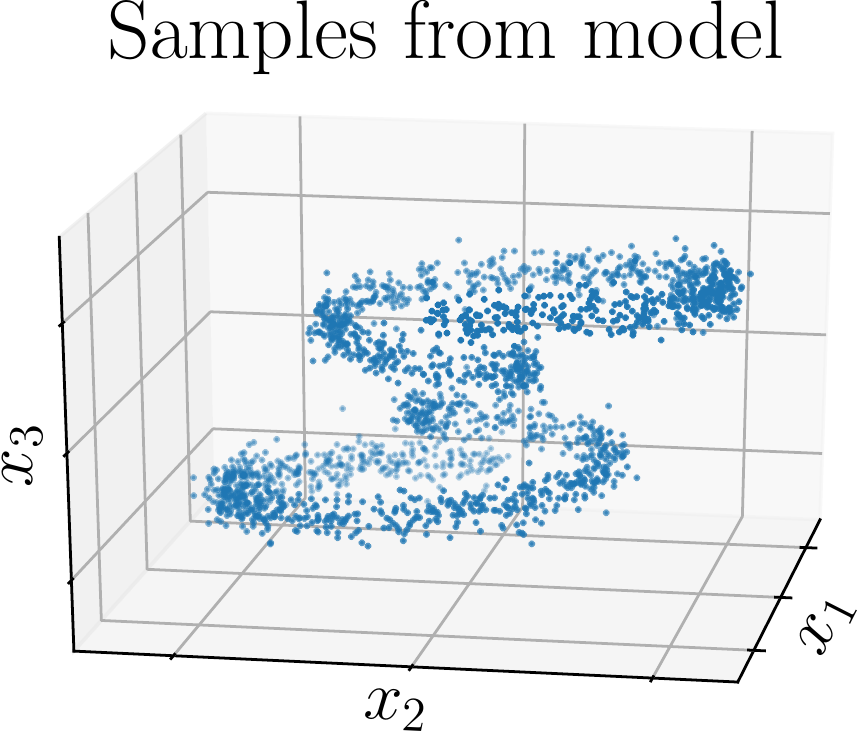}
      \label{}
    \end{subfigure}
    \\[-.1in]
       \begin{subfigure}[c]{\subfigwidth}
      \includegraphics[width=\textwidth]{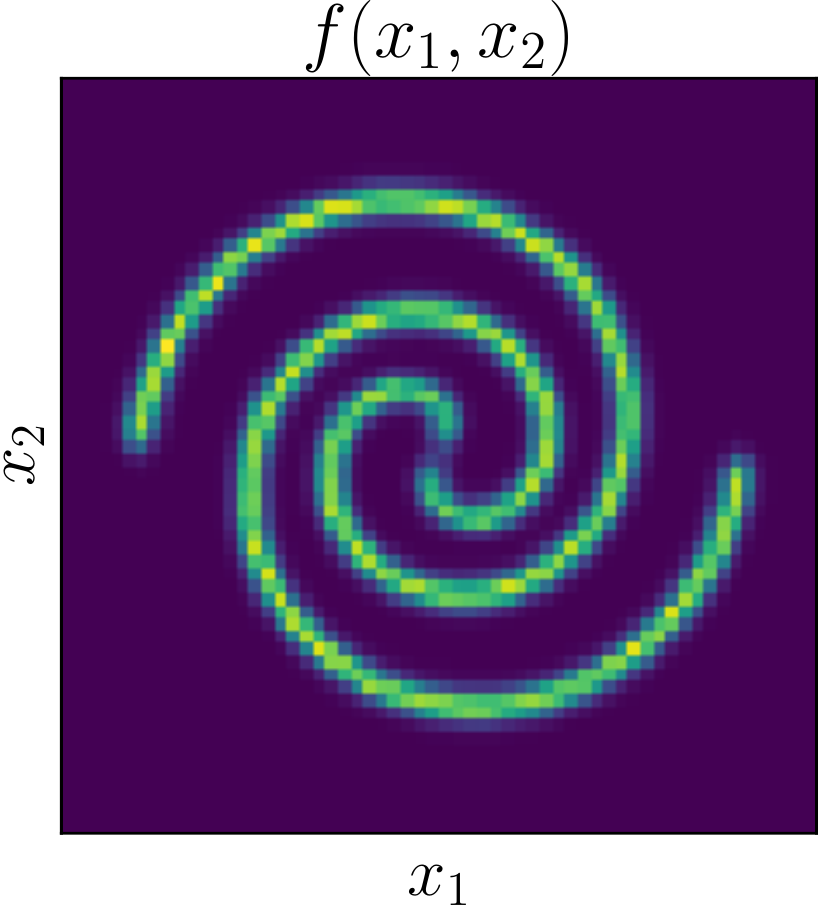}
      \label{}
    \end{subfigure}&
    \begin{subfigure}[c]{\subfigwidth}
      \includegraphics[width=\textwidth]{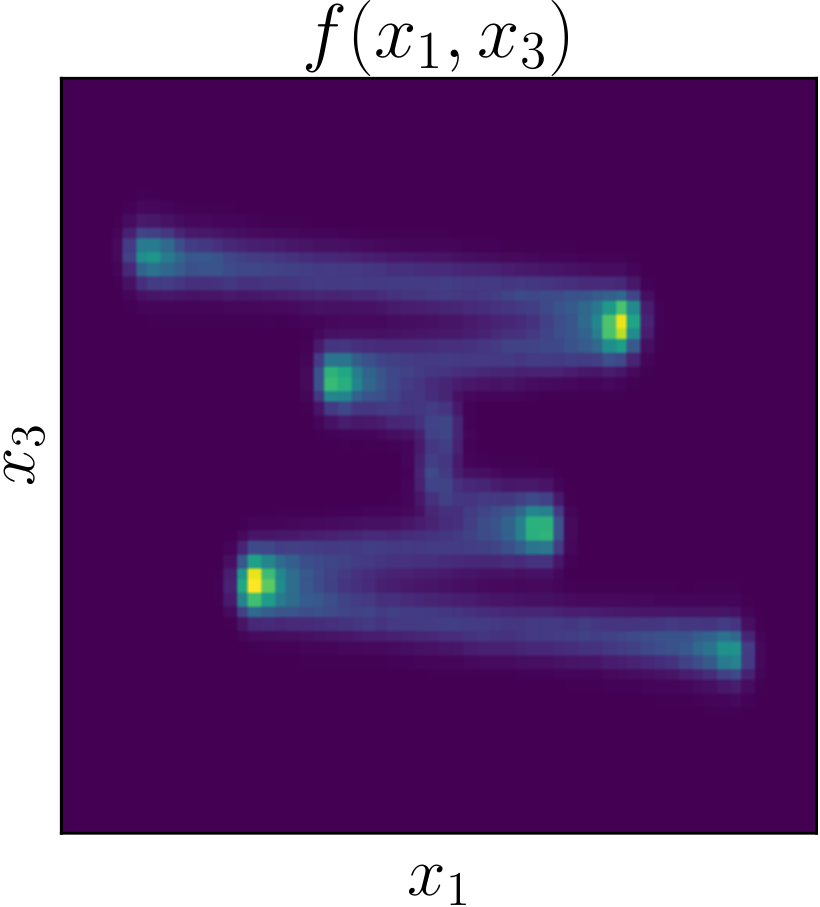}
      \label{}
    \end{subfigure}&
    \begin{subfigure}[c]{\subfigwidth}
      \includegraphics[width=\textwidth]{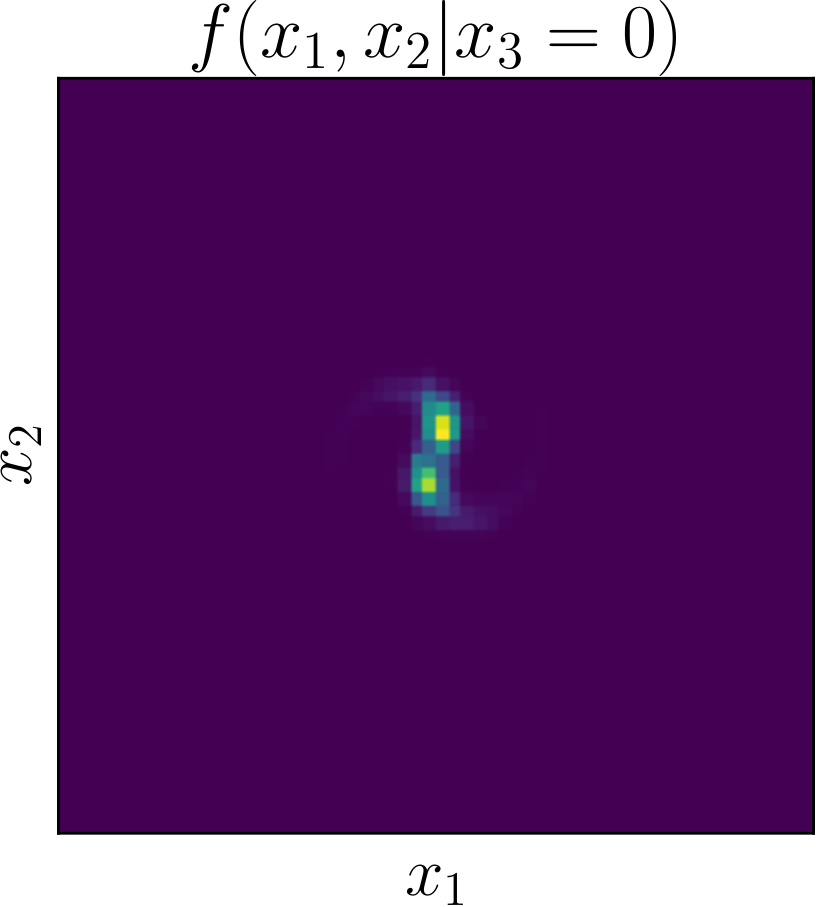}
      \label{}
    \end{subfigure}&
    \begin{subfigure}[c]{\subfigwidth}
      \includegraphics[width=\textwidth]{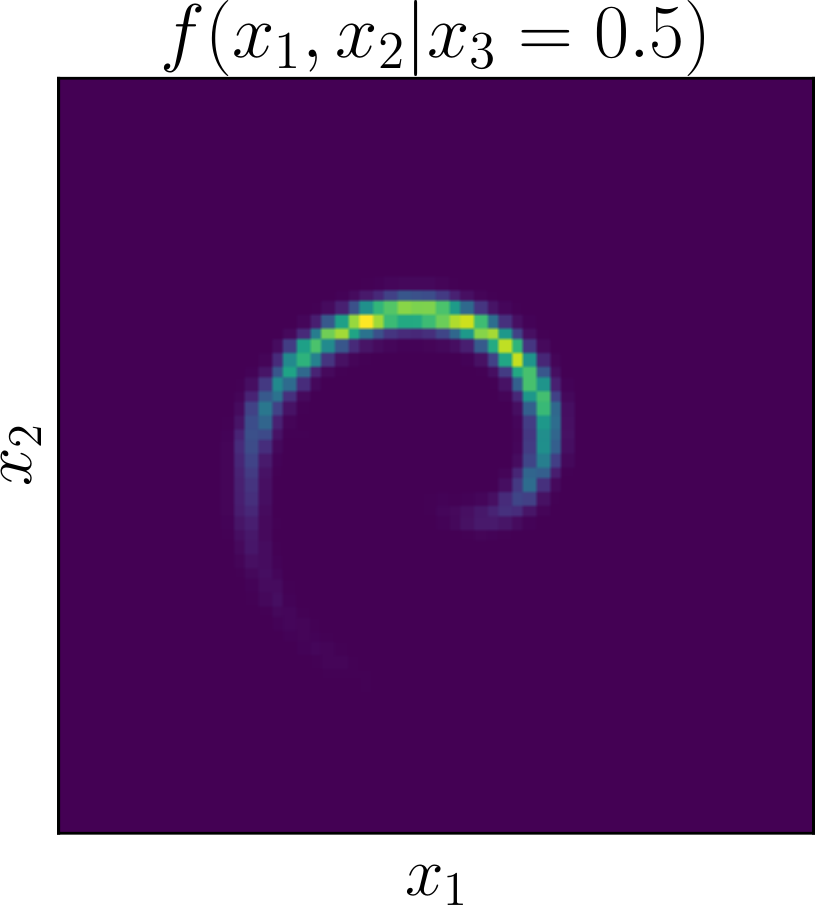}
      \label{}
    \end{subfigure}
        \\[-.1in] 
    \begin{subfigure}[c]{\subfigwidth}
    \vspace{.05in}
      \includegraphics[width=\textwidth]{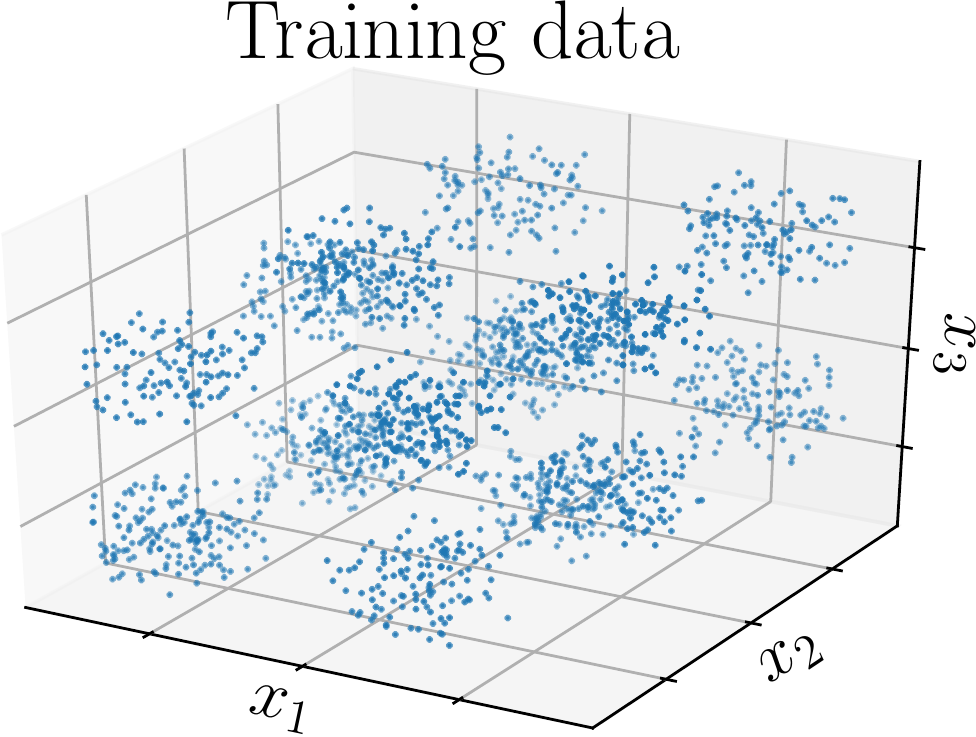}
      \label{}
    \end{subfigure}&
    \begin{subfigure}[c]{\subfigwidth}
      \includegraphics[width=\textwidth]{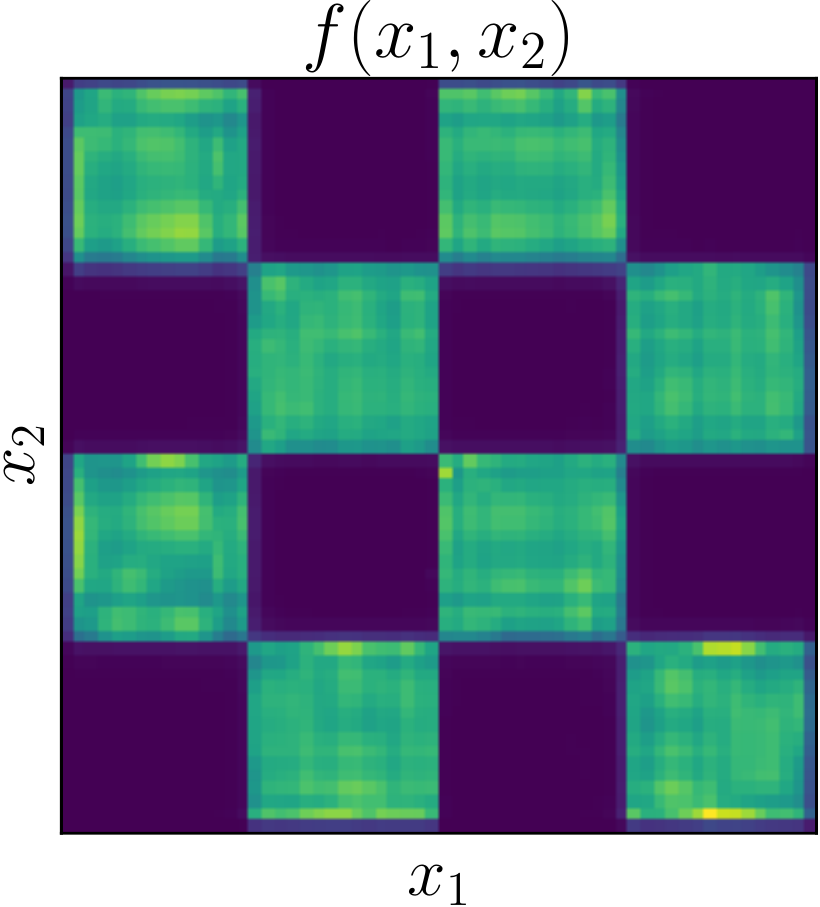}
      \label{}
    \end{subfigure}&
    \begin{subfigure}[c]{\subfigwidth}
      \includegraphics[width=\textwidth]{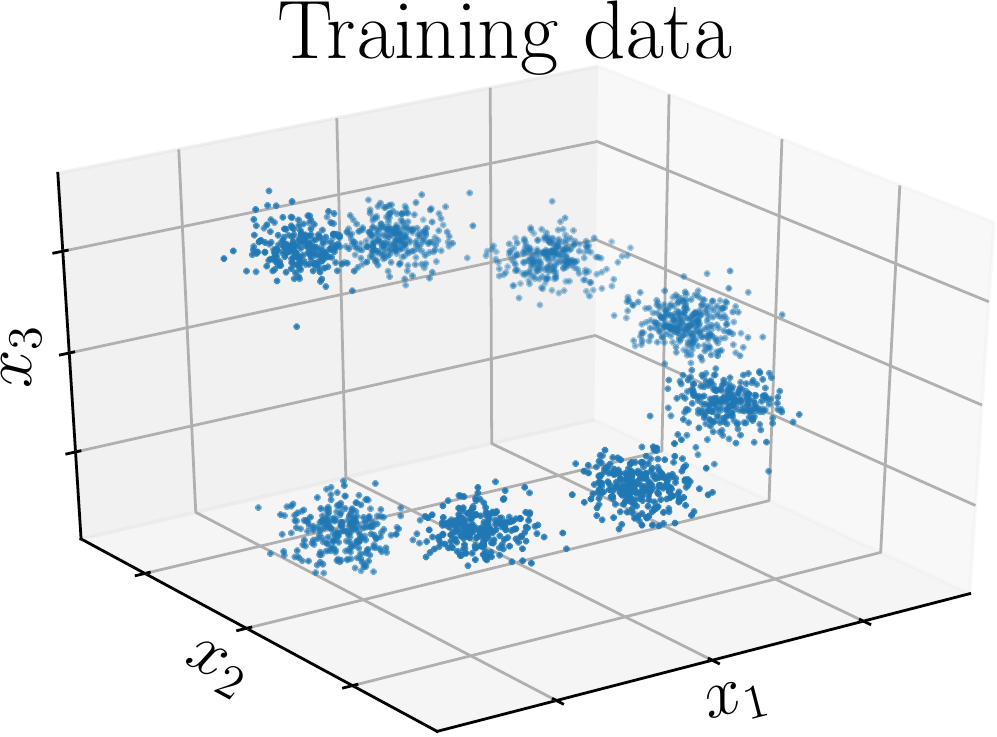}
      \label{}
    \end{subfigure}&
    \begin{subfigure}[c]{\subfigwidth}
      \includegraphics[width=\textwidth]{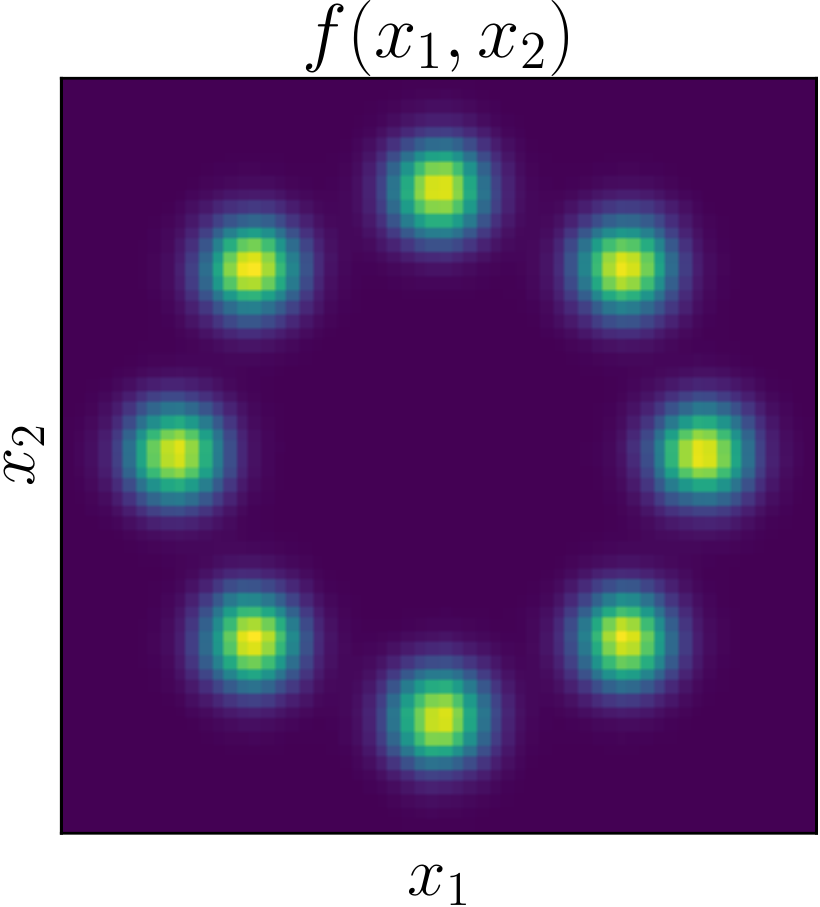}
      \label{}
    \end{subfigure}
  \end{tabular}};
        \draw[black,thick,] (-7,2.3) rectangle (3.43,5.8);
      \node[text width=3cm] at (-1,6) 
    {Training data};
\end{tikzpicture}
\vspace{-.3in}
  \caption{\textbf{Density estimation with closed-form marginals and conditionals.} 
  \textit{First Row:} Panels~1,2: Empirical histograms of training data. Panel 3: Samples from the training data. Panel 4: Samples from the trained MDMA model.
  \textit{Second Row:} Panels 1,2: The marginal density learned by MDMA plotted on a grid. Panels 3,4: Conditional densities learned by MDMA plotted on a grid. 
   \textit{Third Row:} Results on additional datasets: Panels 1,2: Training data and learned marginal density for a 3D checkerboard dataset. Panels 3,4: Similarly for a 3D mixture of Gaussians.
   }
  \label{fig:toy_de}
\end{figure}

We start by considering  3D augmentations of three popular 2D toy probability distributions 
introduced in~\cite{grathwohl2018ffjord}: two spirals, a ring of  8 Gaussians and a checkerboard pattern. These distributions allow to explore the ability of density models
to capture challenging multimodalities and discontinuities~\cite{de2020block, huang2018neural}.
The results, presented in~\Cref{fig:toy_de}, show that MDMA captures all marginal densities with high accuracy, and samples from the learned model appear indistinguishable from the training data. 

%% file: arxiv/exp_mie.tex

\subsection{Mutual information estimation}
\begin{SCfigure}
    \centering
    \includegraphics[height=1.8in]{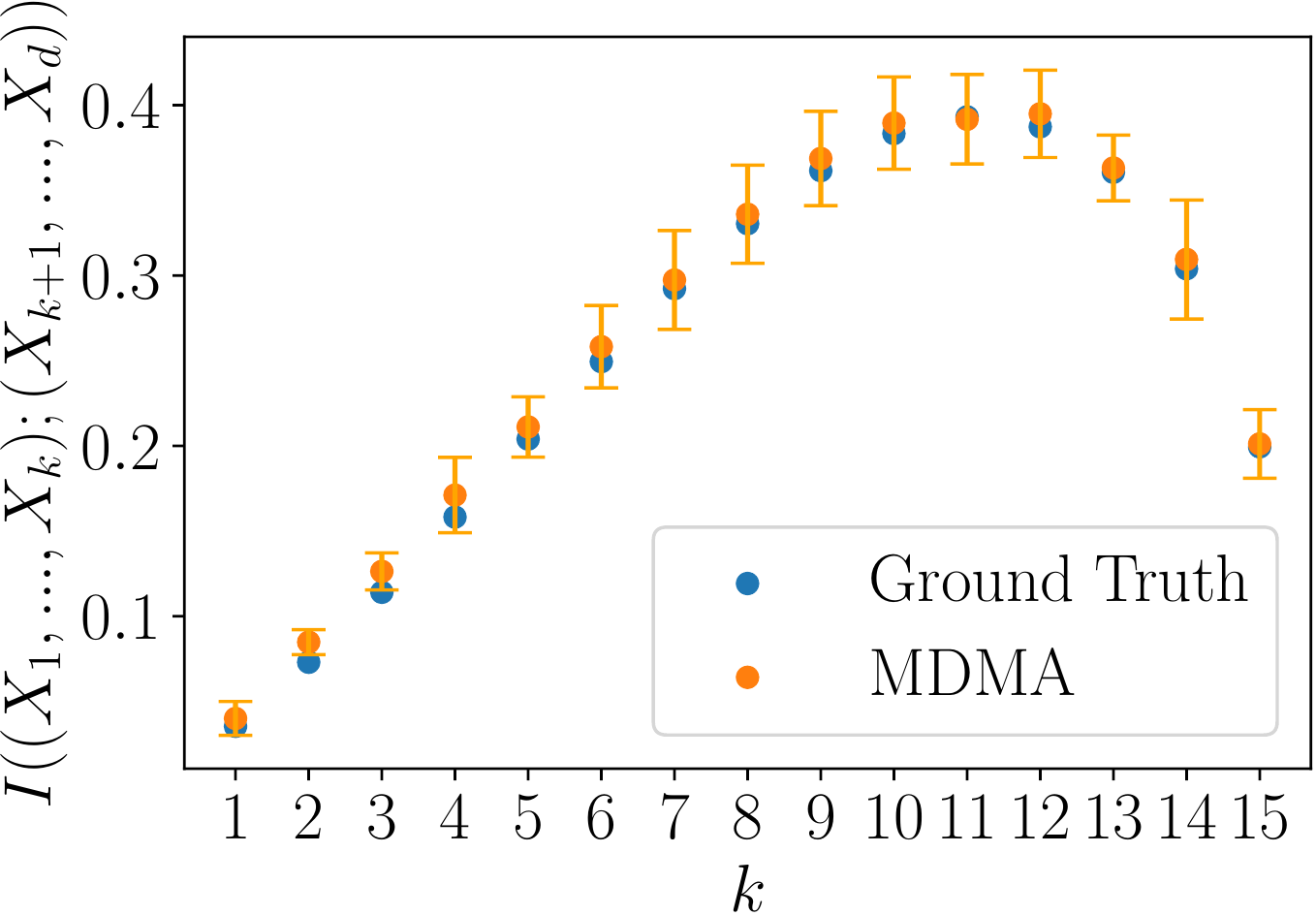}
    \caption{\textbf{Mutual information estimation between subsets of a random vector.}
    We fitted a single MDMA model to samples from a zero-mean $d=16$ Gaussian, with covariance $\Sigma_{ij}=\delta_{ij}+(1-\delta_{ij})(i+j-2)/(5d)$. 
    Monte Carlo estimates of the mutual information~(\ref{eq:MI}) between 
     $(X_1, \dots, X_k)$ and $(X_{k+1}, \dots, X_d)$ for any $k=1,\dots, d-1$ are easily obtained and match closely the exact values.       For each $k$ we average over 5 repetitions of drawing the dataset and fitting. 
    }
    \label{fig:MI_est}
    \vspace{-.5in}
\end{SCfigure}

Given a multivariate probability distribution over some variables $X=(X_1 \dots ,X_d)$, estimating the mutual information 
\begin{equation}
    I(Y;Z)=\int dp_{X}(x)\log\left(\frac{p_{X}(x)}{p_{Y}(y)p_{Z}(z)}\right),
    \label{eq:MI}
\end{equation}
where $Y,Z$ are random vectors defined by disjoint subsets of the $X_i$, requires evaluating $p_{Y}, p_{Z}$ which are marginal densities of $X$. Typically, $Y$ and $Z$ must be fixed in advance, yet in some cases it is beneficial to be able to flexibly compute mutual information between any two subsets of variables. Estimating both $I(Y,Z)$ and $I(Y',Z')$ may be highly inefficient, e.g. if $Y$ and $Y'$ are highly overlapping subsets of $X$. 
Using MDMA however, we can fit a single model for the joint distribution
and easily estimate the mutual information between \textit{any} subset of variables by simply marginalizing over the remaining variables to obtain the required marginal densities. 
Thus a Monte Carlo estimate of (\ref{eq:MI}) can be obtained by evaluating the marginal densities at the points that make up the training set. 
\Cref{fig:MI_est} presents an example of this method, showing the accuracy of the estimates.


%% file: arxiv/exp_demv.tex

\subsection{Density estimation with missing values} \label{sec:demv}

Dealing with missing values in multivariate data is a classical challenge in statistics that has been studied for decades~\cite{little2019}.
The standard solution is the application of a data imputation procedure (i.e., ``filling in the blanks''), 
which requires making structural assumptions. 
In some cases, this is natural, as for the matrix completion problem under a low-rank assumption~\cite{candes2009exact, candes2010matrix}, where the imputed values are the main object of interest.
But the artifacts introduced by data imputation~\cite{beretta2016nearest}
are generally a price that one must 
unwillingly pay in order to perform statistical inference in models that require fully-observed data points.
Two popular, generic techniques for imputation are MICE~\cite{buuren2010mice} and $k$-NN imputation~\cite{troyanskaya2001missing}.
The former imputes missing values by iteratively regressing each missing variable against the remaining variables, while the latter uses averages over the non-missing values at $k$-nearest datapoints.

More formally, let $X \in \mathbb{R}^{d}$ be distributed according to some density $p$ with parameters $\theta$,  let $X_{(0)}, X_{(1)}$ be the non-missing and missing entries of $X$ respectively, and $M\in\{0,1\}^{d}$ a vector indicating the missing entries. 
In the missing-at-random setting (i.e. $M$ is independent of $X_{(1)}$), 
likelihood-based inference using the full likelihood of the model is equivalent to inference using the marginal likelihood~\cite{little2019} $L(X_{(0)}|\theta)=\int p(X|\theta)dX_{(1)}$.
Standard neural network-based density estimators 
must resort to data imputation because
of the impossibility of computing this marginal likelihood.
MDMA however can directly maximize the marginal likelihood for any pattern of missing data at the same (actually slightly cheaper) computational cost as maximizing the full likelihood, without introducing any bias or variance due to imputation.

\begin{figure}
\vspace{-.2in}
  \begin{subfigure}[b]{0.5\textwidth}
    \centering
    \includegraphics[width=\textwidth]{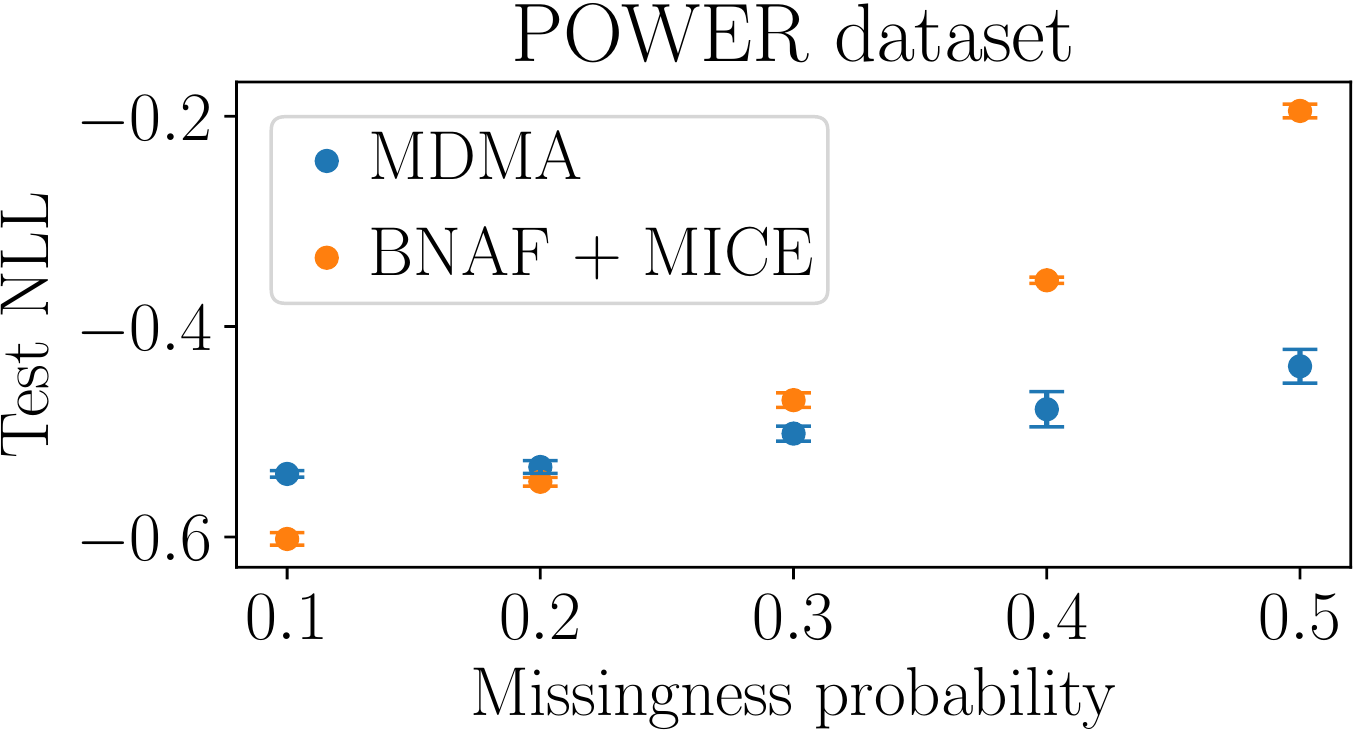}
  \end{subfigure}
    \begin{subfigure}[b]{0.5\textwidth}
    \centering
    \includegraphics[width=\textwidth]{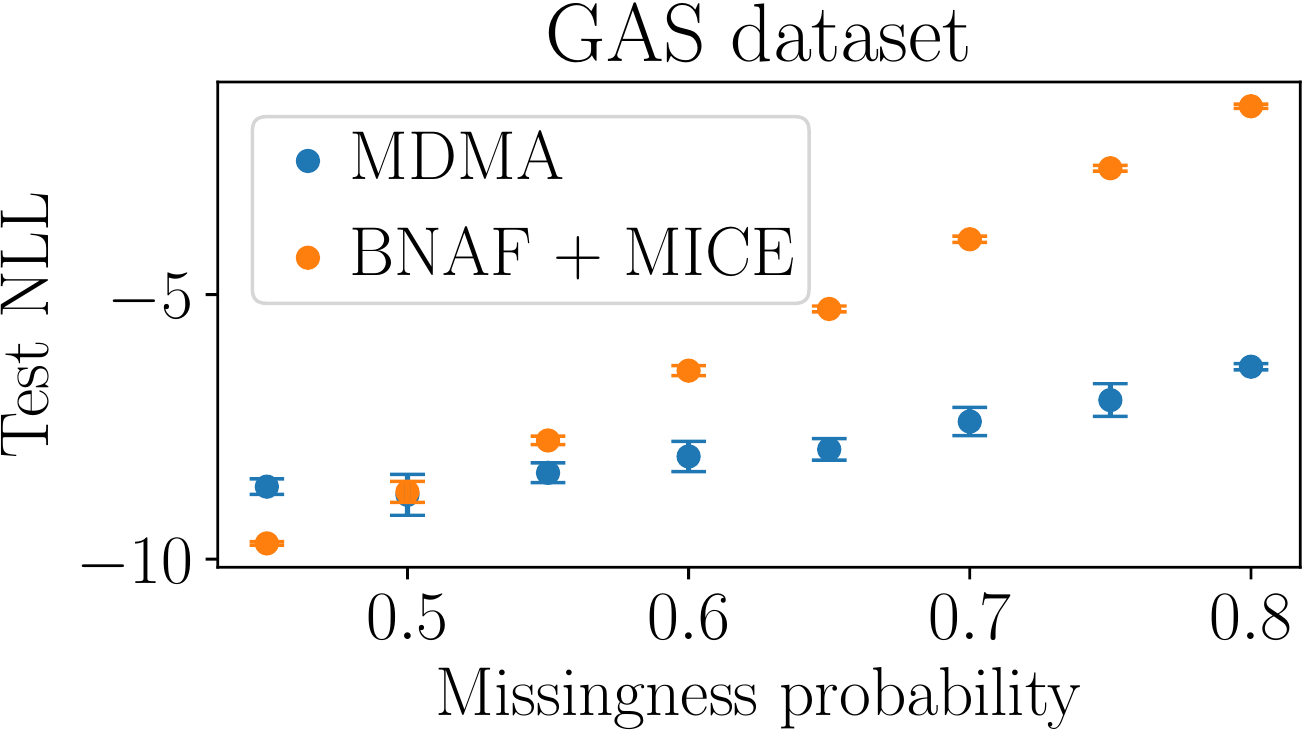}
  \end{subfigure}
    \caption{\textbf{Density estimation with missing data} Test NLL on two density estimation benchmarks, varying the proportion of entries in each data matrix that are designated missing and not used for fitting. We compare MDMA which can fit marginal densities directly with BNAF which achieves state-of-the-art results on the POWER dataset, after performing data imputation using MICE. As the proportion of missing data increases, MDMA outperforms BNAF.}
    \label{fig:missing_data}
\end{figure}

As a demonstration of this capability, we consider the UCI POWER and GAS datasets, following the same pre-processing as \cite{papamakarios2017masked}. We construct a dataset with missing values by setting each entry in the dataset to be missing independently with a fixed probability . We compare MDMA to BNAF~\cite{de2020block}, a neural density model which achieves state-of-the-art results on a number of density estimation benchmarks including GAS. We train MDMA directly on the log marginal likelihood of the missing data, and BNAF by first performing data imputation using MICE \cite{buuren2010mice} and then training using the full log likelihood with the imputed data. The validation loss is the log marginal likelihood for MDMA and the log likelihood of the imputed validation set for BNAF. The test set is left unchanged for both models and does not contain any missing values. We train BNAF using the settings specified in \cite{de2020block} that led to the best performance ($2$ layers and $40d$ hidden units where $d$ is the dimensionality of the dataset). The results are shown in \Cref{fig:missing_data}. 
We find that, as the probability of missingness increases, MDMA significantly outperforms BNAF on both datasets.
Note that, while the proportion of missing values might seem extreme, it is not uncommon in some applications (e.g., proteomics data).
We also trained BNAF using $k$-NN imputation \cite{troyanskaya2001missing}, finding that performance was worse than MICE imputation for all values of $\alpha$.
A comparison of the two methods is provided in \Cref{app:additional_exps}. 

%% file: arxiv/exp_causal.tex

\subsection{Conditional independence testing and causal discovery}

Randomized control trials \cite{Fisher1936} remain the golden standard for causal discovery.
Nonetheless, experiments or interventions are seldom doable, e.g. due to  financial or ethical considerations.
Alternatively, observational data can help uncovering causal relationships \cite{Spirtes2000, Maathuis2015}.
In this context, a class of popular methods targeted at recovering the full causal graph, like PC or FCI \citep{Spirtes2000,strobl2019}, rely on conditional independence (CI) tests.
Letting $X$, $Y$ and $Z$ be random variables, the CI of $X$ and $Y$ given $Z$, denoted $X \independent Y \mid Z$, means that given $Z$, no information about $X$ (or $Y$) can be gained by knowing the value of $Y$ (or $X$).
And testing $H_0\colon X \independent Y \mid Z$ against $H_1\colon X \not \independent Y \mid Z$ is a problem tackled in econometrics \citep{su2007consistent,su2008nonparametric}, statistics \citep{huang2010testing,shah2020hardness}, and machine learning \citep{zhang2011,petersen2021testing}.

Following \citep{petersen2021testing}, denote $U_1 = F(X\mid Z)$ and $U_2 = F(Y\mid Z)$.
It is clear that $H_0$ implies $U_1 \independent U_2$, although the converse does not hold \citep[see e.g.,][]{spanhel2016partial}.
Nonetheless, $U_1 \not \independent U_2$ implies $H_1$, so a test based on the independence between $U_1$ and $U_2$ can still have power.
While the test from \citep{petersen2021testing} is based on estimating the conditional CDFs through quantile regression, we proceed similarly, albeit using the MDMA as a plugin for the conditional distributions.
Our approach is especially appealing in the context of causal discovery, where algorithms require computing many CI tests to create the graph's skeleton.
Instead of having to regress for every test, MDMA estimates the full joint distribution, and its lower dimensional conditionals are then used for the CI tests.

In \Cref{tab:CD}, we present results on inferring the structure of causal graphs using the PC algorithm \cite{Spirtes2000,kalisch2007estimating,sun2007kernel,tillman2009nonlinear,harris2013pc}.
As a benchmark, we use the vanilla (i.e., Gaussian) CI test, and compare it to the PC algorithm obtained with the CI test from \citep{petersen2021testing}, albeit using the MDMA for the conditional distributions.
Synthetic random directed acyclic graphs (DAGs) along with sigmoidal or polynomial mechanisms linking parents to children are sampled using~\cite{JMLR:v21:19-187}.
Each dataset is $d=10$ dimensional and contains 20,000 observations.
We also compare the two algorithms on data from a protein signaling network with $d=11$ \cite{sachs2005causal} for which the ground truth causality graph is known.
Performance is assessed based on the structural Hamming distance (SHD)~\cite{Tsamardinos2006}, that is the $L^1$ norm of the difference between learned adjacency matrices and the truth, as well as a variant of this metric for directed graphs SHD(D) which also accounts for the direction of the edges.
\Cref{tab:CD} shows averages over 8 runs for each setting.
In all cases, MDMA outperforms the vanilla PC in terms of both metrics.
For the synthetic data, we note the large standard deviations, due in part to the fact that we sample randomly from the space of DAG structures, which has cardinality super-exponential in $d$.
An example of the inferred graphs is presented in \Cref{app:additional_exps}.

\begin{table}[]
\caption{\textbf{MDMA for causal discovery.} Conditional densities from a trained MDMA model can be used for causal discovery by allowing to test for conditional independence between variables. Both on synthetic DAG data and real data from a protein signaling network, MDMA infers the graph structure more accurately than a competing method based on quantile regression \cite{petersen2021testing}. The metrics are the structural Hamming distance for the directed (SHD(D)) and undirected (SHD) graph. \label{tab:CD} }
\begin{tabular}{lcccccc}
\toprule
Model & \multicolumn{2}{c}{Sigmoidal DAG, d=10} &  \multicolumn{2}{c}{Polynomial DAG, d=10} & \multicolumn{2}{c}{Sachs \cite{sachs2005causal}, d=11} 
\\ \hline
      & SHD(D)                  & SHD              & SHD(D)                    & SHD                & SHD(D)            & SHD        \\
Gaussian    & $18.6 \pm 3.0$        & $15.6 \pm 2.7$       & $19.8 \pm 4.1$         & $18.9 \pm 4.2$         & 32             & 27             \\
MDMA        & $\mathbf{15.6} \pm 6.1$        & $\mathbf{12.8} \pm 5.2$       & $\mathbf{17.9} \pm 5.3$         & $\mathbf{15.0} \pm 4.5$         & $\mathbf{30.3} \pm 1.8$ & $\mathbf{25.8} \pm 0.7$
\\
\bottomrule
\end{tabular}
\end{table}


%% file: arxiv/exp_UCI.tex

\subsection{Density estimation on real data}

We trained MDMA/nMDMA and the non-marginalizable variant described in \Cref{sec:nmdma} on a number of standard density estimation benchmarks from the UCI repository,\footnote{http://archive.ics.uci.edu/ml/datasets.php} following the pre-processing described in \cite{papamakarios2017masked}. 
\Cref{tab:UCI} compares test log likelihoods of 
MDMA/nMDMA with several other neural density models.  
We find the performance of MDMA on the lower-dimensional datasets  comparable to state-of-the-art models, while for higher-dimensional datasets it appears to overfit. nMDMA achieves state-of-the-art performance on the POWER ($d=6$) dataset, but at the cost of losing the ability to marginalize or condition over subsets of the variables. The width of MDMA was chosen based on a grid search over $\{500, 1000, 2000, 3000, 4000\}$ for each dataset, and the marginal CDF parameters by a search over $\left\{ \left(l=2,w=3\right),\left(l=4,w=5\right)\right\}$. All models were trained using ADAM with learning rate $0.01$, and results for MDMA and nMDMA are averaged over $3$ runs.  
Additional experimental details are provided in \Cref{app:exp_details}. 

\begin{table}[]
\caption{\textbf{General density estimation.} Test log likelihood for density estimation on UCI datasets. The comparison results are reproduced from \cite{bigdeli2020learning}.
}
\begin{tabular}{lcccc}
\toprule
Model  &   {\small POWER} {\footnotesize [d=6]}  &  {\small GAS} {\footnotesize [d=11]}           & {\small HEPMASS} {\footnotesize [d=21]}   & {\footnotesize MINIBOONE } {\footnotesize [d=43]}        \\ 
\hline
Kingma et al. 2018   \cite{kingma2018}      & $0.17 \pm .01$ & $8.15 \pm .4$            & $-18.92 \pm .08$         & $-11.35 \pm .07$          \\
Grathwohl et al. 2019     \cite{grathwohl2018ffjord}      & $0.46 \pm .01$ & $8.59 \pm .12$           & $-14.92 \pm .08$         & $-10.43 \pm .04$          \\
Huang et al. 2018         \cite{huang2018neural}          & $0.62 \pm .01$ & $11.96 \pm .33$          & $-15.08 \pm .4$          & $-8.86 \pm .15$           \\
Oliva et al. 2018         \cite{oliva2018transformation}  & $0.60 \pm .01$ & $\mathbf{12.06} \pm .02$ & $-13.78 \pm .02$         & $-11.01 \pm .48$          \\
De Cao et al. 2019 \cite{de2020block}                                         & $0.61 \pm .01$ & $\mathbf{12.06} \pm .09$ & $-14.71 \pm .38$         & $-8.95 \pm .07$           \\
Bigdeli et al. 2020 \cite{bigdeli2020learning}                                                  & $0.97 \pm .01$ & $9.73 \pm 1.14$          & $\mathbf{-11.3} \pm .16$ & $\mathbf{-6.94} \pm 1.81$ \\ \hline
MDMA       &       $0.57 \pm .01$                                   &       $8.92 \pm 0.11$         &      $-20.8 \pm .06$                                 &      $-29.0 \pm .06$                     \\
nMDMA                                            &           $\mathbf{1.78} \pm .12$         &   $8.43 \pm .04$                       &         $-18.0 \pm 0.91$                 &  $-18.6 \pm .47$    
\\

\bottomrule
\end{tabular}
\label{tab:UCI}
\vspace{-.1in}
\end{table}

%% file: arxiv/discussion.tex

\section{Discussion}
\label{sec:discussion}
MDMAs offer the ability to obtain, from a single model, closed form probabilities, marginals and conditionals for any subset of the variables. These properties enable one to straightforwardly use the model to solve a diverse array of problems, of which we have demonstrated only a few: mutual information estimation between arbitrary subsets of variables, inference with missing values, and conditional independence testing targeted at multivariate causal discovery. 
In addition to these, MDMA's marginalization property can be used for anomaly detection with missing values~\cite{dietterich2018anomaly}. We have shown that MDMA can fit data with missing values without requiring imputation, yet if one is interested in data imputation for downstream tasks, the ability to sample from arbitrary conditional distributions means that MDMA can be used for imputation as well. 

Additionally, in some application areas (e.g., financial risk management), powerful models exist for the univariate distributions, and marginal distributions are then glued together using copulas~\cite{mcneil2015quantitative}.
However, popular copula estimators suffer from the same drawbacks as modern neural network density estimators with regard to marginalization and conditioning. Using MDMA for copula estimation (say by replacing the kernel density estimator by MDMA in the formulation of \cite{Geenens2017-tg}), one can then obtain copula estimators that do not suffer from these deficiencies. 

The main shortcoming of MDMA is the linearity in the combination of the products of univariate CDFs which appears to limit the expressivity of the model. The study of tensor decompositions is an active area of research, and novel constructions, ideally adapted specifically for this task, could lead to improvements in this regard despite the linear structure. 

%% file: arxiv/ack.tex
\section*{Acknowledgements}

The work of DG is supported by a Swartz fellowship. 
The work of AP is supported by the Simons Foundation, the DARPA NESD program, NSF NeuroNex Award DBI1707398 and The Gatsby Charitable Foundation.

%% file: arxiv/proofs.tex

\section{Proofs} \label{app:proofs}

\subsection{Proof of \Cref{prop:HT_sampling}}

The proof follows directly from Algorithm \ref{algo:ht_sampling}. The distribution $F_{A^{\mathrm{HT}},\Phi})=\left\langle A^{\mathrm{HT}},\Phi\right\rangle$ is a mixture model, and thus in order to sample from it we can first draw a single mixture component (which is a product of univariate CDFs) and then sample from this single component. The mixture weights are the elements of the tensor $A^{\mathrm{HT}}$ given by the diagonal HT decomposition \cref{eq:HT_mixture}. In the next section, we add details on the sampling process for the sake of clarity.

\subsubsection{Details on the sampling for the HT model}
Define a collection of independent categorical variables $R=\{R^\ell_{i,j}\}$ taking values in $[m]$, where $\ell \in [p], i \in [m]$ and for any $\ell$, $j \in [2^{p-\ell}]$. These variables are distributed according to  
\begin{align*}
	\forall\ell,i,j:\quad\mathbb{P}\left[R_{i,j}^{\ell}=k\right]=\lambda_{i,k,j}^{\ell},
\end{align*}
where $\{ \lambda^{\ell} \}_{\ell=1}^{p}$ are the parameters of the HT decomposition. The fact that the parameters are nonnegative and $\sum_{k=1}^m \lambda_{k,i,j}^{\ell} = 1$ ensures the validity of this distribution.

With the convention $R_{k_{p+1,1},1}^{p}=R_{1,1}^{p}$, define the event
\begin{align*}
\underset{\ell=1}{\overset{p}{\bigcap}}\left(\cap_{j_{\ell}=1}^{2^{p-\ell}}\left\{ R_{k_{\ell+1,\left\lceil j_{\ell}/2\right\rceil },j_{\ell}}^{\ell}=k_{\ell,j_{\ell}}\right\} \right) &=
    \left\{ R_{1,1}^{p}=k_{p,1}\right\} \\
&\phantom{=}\bigcap\left(\cap_{j=1}^{2}\left\{ R_{k_{p,1},j}^{p-1}=k_{p-1,j}\right\} \right)\\
&\phantom{=}\,\,\,\,\vdots\\
&\phantom{=}\bigcap\left(\cap_{j=1}^{d/2}\left\{ R_{k_{2,\left\lceil j/2\right\rceil },j}^{1}=k_{1,j}\right\} \right).
\end{align*}
Let $(\tilde{X}_{1},\dots,\tilde{X}_{d}, R)$ be a random vector such that
\begin{align} \label{eq:sample_HT}
\mathbb{P}\left[\tilde{X}_{1}\leq x_{1},\dots,\tilde{X}_{d}\leq x_{d}\left|\underset{\ell=1}{\overset{p}{\bigcap}}\left(\cap_{j_{\ell}=1}^{2^{p-\ell}}\left\{ R_{k_{\ell+1,\left\lceil j_{\ell}/2\right\rceil },j_{\ell}}^{\ell}=k_{\ell,j_{\ell}}\right\} \right)\right.\right] 
= \prod_{i=1}^{d}\varphi_{k_{1,\left\lceil i/2\right\rceil },i}(x_{i}),
\end{align}
which implies that the distribution of $(\tilde{X}_{1},\dots,\tilde{X}_{d})$ obtained after conditioning on a subset of the $\{R^\ell_{i,j}\}$ in this way is equal to a single mixture component in $F_{\mathrm{HT}}=\left\langle A,\Phi\right\rangle$.
Thus, based on a sample of $R$, one can sample $\tilde{X}_{i}$ by inverting the univariate CDFs $\varphi_{k_{1,\left\lceil i/2\right\rceil },i}$ numerically and parallelizing over $i$. 
Numerical inversion is trivial since the functions are increasing and continuously differentiable, and this can be done for instance using the bisection method.
It remains to sample a mixture component.

Assume that a sample $\{R_{i,j}^{\ell}\}$ for a sequence of variables as in \cref{eq:sample_HT} is obtained e.g. from Algorithm \ref{algo:ht_sampling}.
With the convention $\lambda_{k_{p+1,1},k,1}^{p}=\lambda_{k}^{p}$, since 
\begin{align*}
\mathbb{P}\left[\underset{\ell=1}{\overset{p}{\bigcap}}\left(\cap_{j_{\ell}=1}^{2^{p-\ell}}\left\{ R_{k_{\ell+1,\left\lceil j_{\ell}/2\right\rceil },j_{\ell}}^{\ell}=k_{\ell,j_{\ell}}\right\} \right)\right]=\prod\limits _{\ell=1}^{p}\prod\limits _{j_{\ell}=1}^{2^{p-\ell}}\lambda_{k_{\ell+1,\left\lceil j_{\ell}/2\right\rceil },k_{\ell,j_{\ell}},j_{\ell}}^{\ell},
\end{align*}
sampling from the categorical variables in this fashion is equivalent to sampling a mixture component.
It follows that by first sampling a single mixture component and then sampling from this component, one obtains a sample from $F_{\mathrm{HT}}$. 

The main loop in Algorithm \ref{algo:ht_sampling} samples such a mixture component, and there are $p= \log_2 d $ layers in the decomposition, so the time complexity of the main loop is $O(\log d)$, and aside from storing the decomposition itself this sampling procedure requires storing only $O(d)$ integers. This logarithmic dependence is only in sampling from the categorical variables which is computationally cheap. This not only avoids the linear time complexity common in sampling from autoregressive models (without using distillation), but the space complexity is also essentially independent of $m$ since only a single mixture component is evaluated per sample. 



\subsection{Proof of \Cref{prop:universal_1}}



Assume that the activation function $\sigma$ is increasing, continuously differentiable, and such that $\lim_{x\to-\infty}\sigma(x) = 0$ and $\lim_{x\to\infty} \sigma(x) = 1$.
\Cref{prop:universal_1} then follows immediately from \Cref{prop:universal_1a} and the fact that $\cup_r \Phi_{1,r,\sigma} \subseteq \cup_{l,r} \Phi_{l,r,\sigma}$.
\begin{remark}
    In practice, we use the activation $\sigma(x) = x + a \tanh(x)$ for some $a > -1$.
    While it does not satisfy the assumptions, the arguments in the proof of \Cref{prop:universal_1b} can be modified in a straightforward manner to cover this activation (see \Cref{rmk:activation_cdf}).
\end{remark}
\begin{proposition}\label{prop:universal_1a}
$\cup_r \Phi_{1,r,\sigma}$ is dense in $\Fcal_1$ with respect to the uniform norm.
\end{proposition}
Letting
\begin{align*}
 \widetilde{\Fcal}_1 &= \{\widetilde{F}\colon \R \to \R,\, \widetilde{F}(x) = \log F(x)/(1-F(x)),\, F \in \Fcal_1\},\\
\widetilde{\Phi}_{l,r,\sigma} &= \{\widetilde{\varphi} \colon \R \to \R,\, \widetilde{\varphi}(x) = \log \varphi(x)/(1-\varphi(x)),\,\varphi \in \Phi_{l,r,\sigma}\},
\end{align*}
the proof of \Cref{prop:universal_1a} relies on the following proposition.
\begin{proposition}\label{prop:universal_1b}
$\cup_r \widetilde{\Phi}_{1,r,\sigma}$ is dense in $\widetilde{\Fcal}_1$ with respect to the uniform norm.
\end{proposition}

\subsubsection{Proof of \Cref{prop:universal_1a}}
This proof is similar to that of \cite[][Theorem 2]{dugas2009incorporating}, which deals with functions with positive outputs.
We want to show that, for any $F \in \Fcal_1$, compact $K \subset \R$, and $\epsilon > 0$, there exists $\varphi \in \cup_r \Phi_{1,r,\sigma}$ such that
\begin{align*}
     \| \varphi - F \|_{\infty,K}   = \sup_{x \in K} |\varphi(x) - F(x) | \le \epsilon.
\end{align*}

Denote the sigmoid function by $\rho(x) = 1/(1 + e^{-x})$ and define the function $\widetilde{F}\colon \R \to \R$ by $\widetilde{F}(x) = \log F(x)/(1-F(x))$, so that $F = \rho \circ \widetilde{F}$.
By \Cref{prop:universal_1b},
there exists $ \widetilde{\varphi} \in \cup_{r} \widetilde{\Phi}_{1,r,\sigma}$ such that
\begin{align*}
\sup_{x \in K} |\widetilde{\varphi}(x) - \widetilde{F}(x)| \le 4 \epsilon.
\end{align*}
Thus, letting $\varphi = \rho \circ \widetilde{\varphi}$, we have
\begin{align*}
    |\varphi(x) - F(x)|  = |\rho \circ \widetilde{\varphi}(x) - \rho \circ \widetilde{F}(x)|   \le \sup_{x \in K}\rho(x)'|\widetilde{\varphi}(x) - \widetilde{F}(x)|   \le \epsilon.
\end{align*}
Since $\Phi_{1,r,\sigma} = \{\rho \circ \widetilde{\varphi}\colon \widetilde{\varphi} \in  \widetilde{\Phi}_{1,r,\sigma}\}$, the result follows.

\subsubsection{Proof of \Cref{prop:universal_1b}}
This proof is similar to that of \cite[][Theorem 3.1]{Daniels2010}, which is incomplete and only deals with the sigmoid activation.
First, note that $ \widetilde{\Fcal}_1$ is the space of strictly increasing and continuously differentiable functions.
Therefore, for any $\widetilde{F} \in \widetilde{\Fcal}_1 $ and interval $K = [K_1, K_2]$, we can write, for any $x \in K$,
\begin{align*}
    \widetilde{F}(x) 
    =\widetilde{F}(K_1) + \int_{\widetilde{F}(K_1)}^{\widetilde{F}(K_2)} \Ind_{\widetilde{F}(x) \ge u} du =\widetilde{F}(K_1) + \int_{\widetilde{F}(K_1)}^{\widetilde{F}(K_2)} \Ind_{x \ge \widetilde{F}^{-1}(u)}du,
\end{align*}
where the existence of the inverse $\widetilde{F}^{-1}$ is guaranteed by the fact that $\widetilde{F}$ is strictly increasing and continuous.
Thus, for $\widetilde{F}(K_1) = u_0 < u_1 < \cdots < u_k = \widetilde{F}(K_2)$ a partition of $[\widetilde{F}(K_1),\widetilde{F}(K_2)]$ with $u_{j+1} - u_j \le \epsilon /2(\widetilde{F}(K_2) - \widetilde{F}(K_1))$, $x_j = \widetilde{F}^{-1}(u_j)$ and
\begin{align*}
  G(x) = \widetilde{F}(K_1) + \sum_{j=1}^k \Ind_{x \ge x_j} (u_{j} - u_{j-1}),
\end{align*}
we have $| \widetilde{F}(x) - G(x)| \le \epsilon/2$, namely the approximation error of the Riemann sum for increasing functions.
Let $a > 0$ and $\widetilde{\varphi} \in \widetilde{\Phi}_{1,k,\sigma}$ obtained by setting $b_1 =  \widetilde{F}(K_1)$, as well as $(W_1)_{1,j} = (u_{j} - u_{j-1})/a > 0$, $(b_{0})_j =-ax_j$ and $(W_0)_{j,1} = a$ for $1 \le j \le k$, then
\begin{align*}
   |G(x) - \widetilde{\varphi}(x)| \le \sum_{j=1}^k (u_{j+1} - u_j) \left|\Ind_{x \ge x_j} - \sigma(a(x - x_j))\right|.
\end{align*}
By the assumptions on $\sigma$, it is clear that $|\Ind_{x\ge0} - \sigma(ax)|$ can be made arbitrarily small.
Thus, taking $a$ large enough so that $  |G(x) - \widetilde{\varphi}(x)| \le \epsilon/2$, we have
\begin{align*}
    | \widetilde{F}(x) - \widetilde{\varphi}(x)| \le | \widetilde{F}(x) - G(x)| +|G(x) - \widetilde{\varphi}(x)| \le  \epsilon.
\end{align*}

\begin{remark}\label{rmk:activation_cdf}
   Let $\sigma(x) = x + a \tanh(x)$ for some $a > -1$ and $\widetilde{\varphi} \in \widetilde{\Phi}_{1,k,\sigma}$ obtained by setting $b_1 =  \widetilde{F}(K_1) + 1/2$, as well as $(W_1)_{1,j} = (u_{j} - u_{j-1})/2a > 0$, $(b_{0})_j =-ax_j$ and $(W_0)_{j,1} = |a|$ for $1 \le j \le k$, then
\begin{align*}
   |G(x) - \widetilde{\varphi}(x)| 
   &\le\sum_{j=1}^k (u_{j+1} - u_j) \left|\Ind_{x \ge x_j} - \tanh(|a|(x - x_j))/2 - 1/2\right| \\
   &\phantom{=}+  \sum_{j=1}^k (u_{j+1} - u_j) |x - x_j|/2|a|.
\end{align*}
Because $a$ is arbitrary, one can take it large enough so that  $  |G(x) - \widetilde{\varphi}(x)| \le \epsilon/2$ as above.
\end{remark}


\subsection{Proof of \Cref{prop:universal_d}}

Consider the classes of order $d$ tensored-valued functions with $m$ dimensions per mode defined as
\begin{align*}
    \Phi_{m,d,l,r,\sigma} &= \{ \Phi \colon \R^d \times [m]^d \to [0,1], \, \Phi(\bm x)_{i_1, \dots, i_d} = \textstyle \prod\limits_{j=1}^{d}\varphi_{i_{j},j}(x_{j}),\, \varphi_{i,j} \in \Phi_{l,r,\sigma}\}, \\
    \Fcal_{m,d} &= \{ \Phi \colon \R^d \times [m]^d \to [0,1], \, \Phi(\bm x)_{i_1, \dots, i_d} = \textstyle \prod\limits_{j=1}^{d}F_{i_{j},j}(x_{j}),\, F_{i,j} \in \Fcal_1 \}
\end{align*}
as well as the class of neural network-based and $\Fcal_1$-based MDMAs, that is
\begin{align*}
    \textrm{MDMA}_{m,d,l,r,\sigma} &= \{ F_{A,\Phi}\colon\R^d \to [0,1],\,F_{A,\Phi}(\bm x)= \langle A, \Phi(\bm x) \rangle, \, A \in \Acal_{d,m},\, \Phi \in  \Phi_{m,d,l,r,\sigma} \}, \\
 \textrm{MDMA}_{m,d,\Fcal_1} &=  \{F_{A,\Phi} \colon\R^d \to [0,1],\, F_{A,\Phi}(\bm x)= \langle A, \Phi(\bm x) \rangle, \, A \in \Acal_{d,m},\, \Phi \in  \Fcal_{m,d}\}.
\end{align*}
We can now state the following proposition.
\begin{proposition}\label{prop:universal_da}
$\cup_{l,r} \mathrm{MDMA}_{m,d,l,r,\sigma}$ is dense in $ \mathrm{MDMA}_{m,d,\Fcal_1}$ with respect to the uniform norm
\end{proposition}
\Cref{prop:universal_d} then follows immediately from the fact that $\cup_{m} \textrm{MDMA}_{m,d,\Fcal_1}$ is the space of multivariate mixture distributions admitting a density, which is dense in $\Fcal_d$ with respect to the uniform norm (see e.g.,  \cite[][Theorem 33.2]{DasGupta2008}, \cite[][Theorem 5]{Cheney2009}, or \cite[][Corollary
11]{Nguyen2019}).

\subsubsection{Proof of \Cref{prop:universal_da}}

With $A \in \Acal_{d,m}$, $\Phi_1 \in \Phi_{m,d,l,r,\sigma}$,  and $\epsilon > 0$ and a compact $K = K_1 \times \cdots \times K_d \subset \R^d$, we want to prove that there exists
$A_2 \in \Acal_{d,m}$ and $\Phi_2 \in \Fcal_{m,d}$, such that $\sup_{\bm x \in K} |F_{A,\Phi_1}(\bm x) - F_{A_2,\Phi_2}(\bm x)| \le \epsilon$.
Assuming that we can show $\sup_{\bm x \in K} |\Phi_1(\bm x)_{i_1,\dots,i_d}-  \Phi_2(\bm x)_{i_1,\dots,i_d}| \le \epsilon$, the result would then follow from setting $A_2 = A$ and the fact that $F_{A,\Phi_1}(\bm x) - F_{A,\Phi_2}(\bm x) = \langle A, \Phi_1(\bm x) -  \Phi_2(\bm x) \rangle $ implies
\begin{align*}
    \sup_{\bm x \in K} | F_{A,\Phi_1}(\bm x) - F_{A,\Phi_2}(\bm x) | = 
    &\le \sum_{i_1,\dots,i_d} A_{i_1,\dots,i_d} \sup_{\bm x \in K} |\Phi_1(\bm x)_{i_1,\dots,i_d}-  \Phi_2(\bm x)_{i_1,\dots,i_d}|
    = \epsilon.
\end{align*}

With $\delta = \epsilon^{1/d}$, by \Cref{prop:universal_1}, there exists $l$, $w$, and $\{\varphi\}_{i \in [m],j \in [d]}$ with $\varphi_{i,j} \in  \Phi_{l,r,\sigma}$, such that
\begin{align*}
    \max_{i \in [m],j \in [d]} \sup_{x_j \in K_j}|\varphi_{i,j}(x_j) - F_{i,j}(x_j)| \le \delta.
\end{align*}
Thus, we have that
\begin{align*}
   |\Phi_1(\bm x)_{i_1,\dots,i_d}-  \Phi_2(\bm x)_{i_1,\dots,i_d}| = 
    | \prod\limits_{j=1}^{d}\varphi_{i_j,j}(x_{j}) - \prod_{j=1}^d F_{i,j}(x_j)| \le \delta^d = \epsilon.
\end{align*}

%% file: arxiv/exp_toy_checkerboard.tex

\begin{figure}[h]
  \centering
  \renewcommand{\tabcolsep}{1pt}
  \begin{tikzpicture}
    \node {\begin{tabular}[c]{cccc}
\begin{subfigure}[c]{\subfigwidth}
      \includegraphics[width=\textwidth]{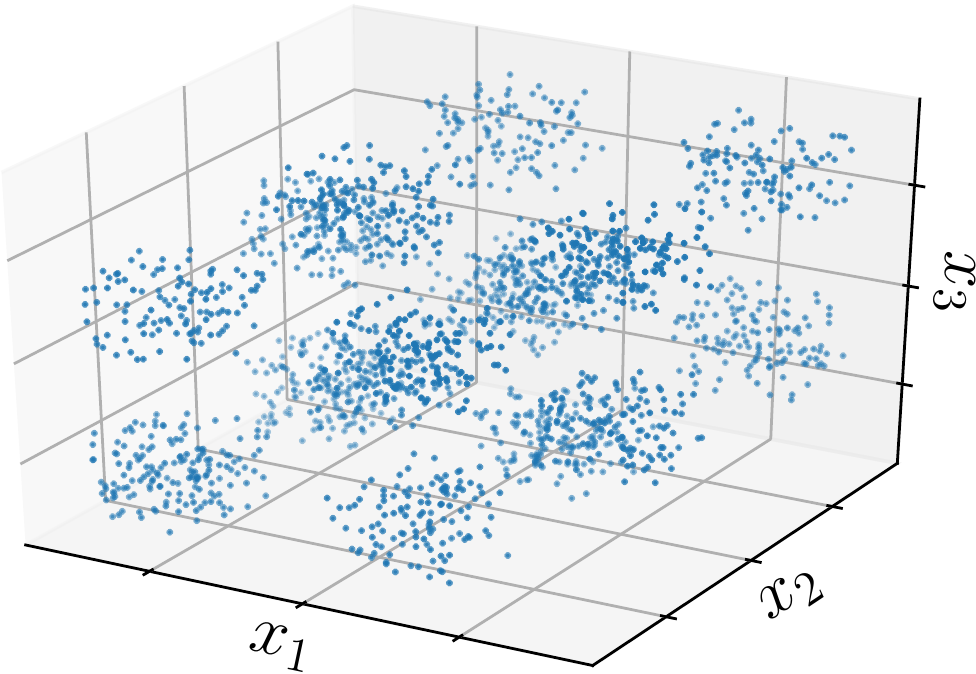}
      \label{}
    \end{subfigure}&
    \begin{subfigure}[c]{\subfigwidth}
      \includegraphics[width=\textwidth]{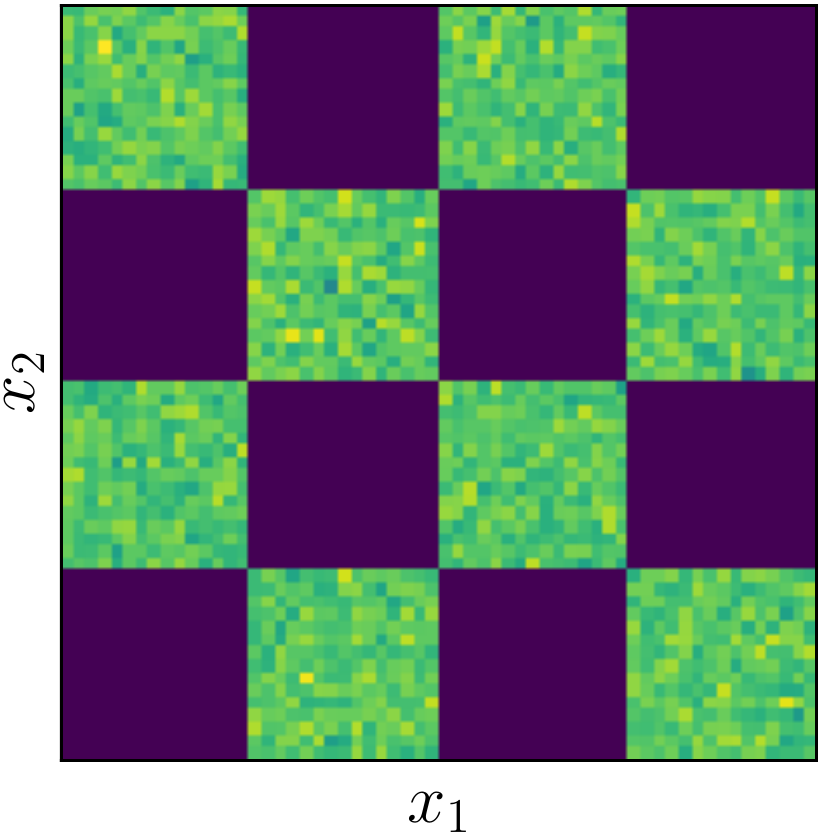}
      \label{}
    \end{subfigure}&
    \begin{subfigure}[c]{\subfigwidth}
      \includegraphics[width=\textwidth]{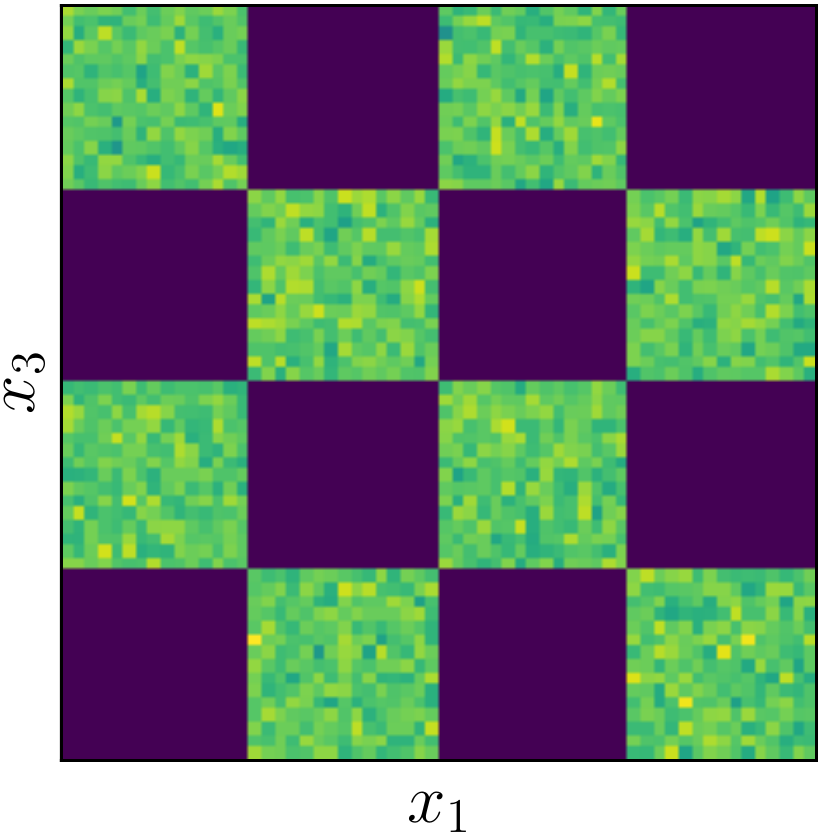}
      \label{}
    \end{subfigure}&
    \begin{subfigure}[c]{\subfigwidth}
      \includegraphics[width=\textwidth]{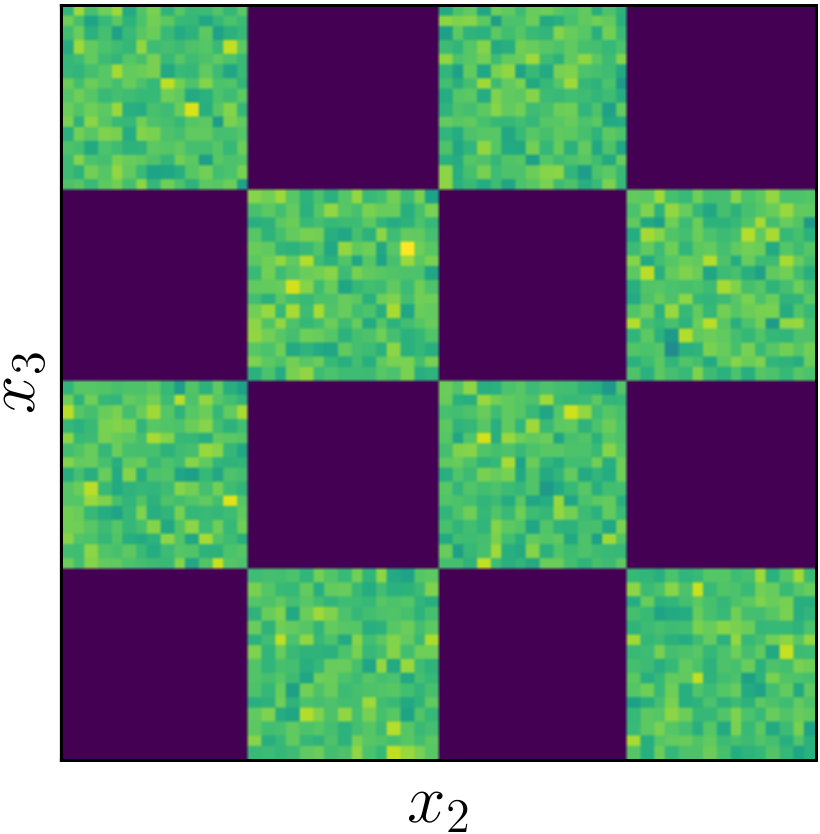}
      \label{}
    \end{subfigure}
    \\[-.1in]
       \begin{subfigure}[c]{\subfigwidth}
      \includegraphics[width=\textwidth]{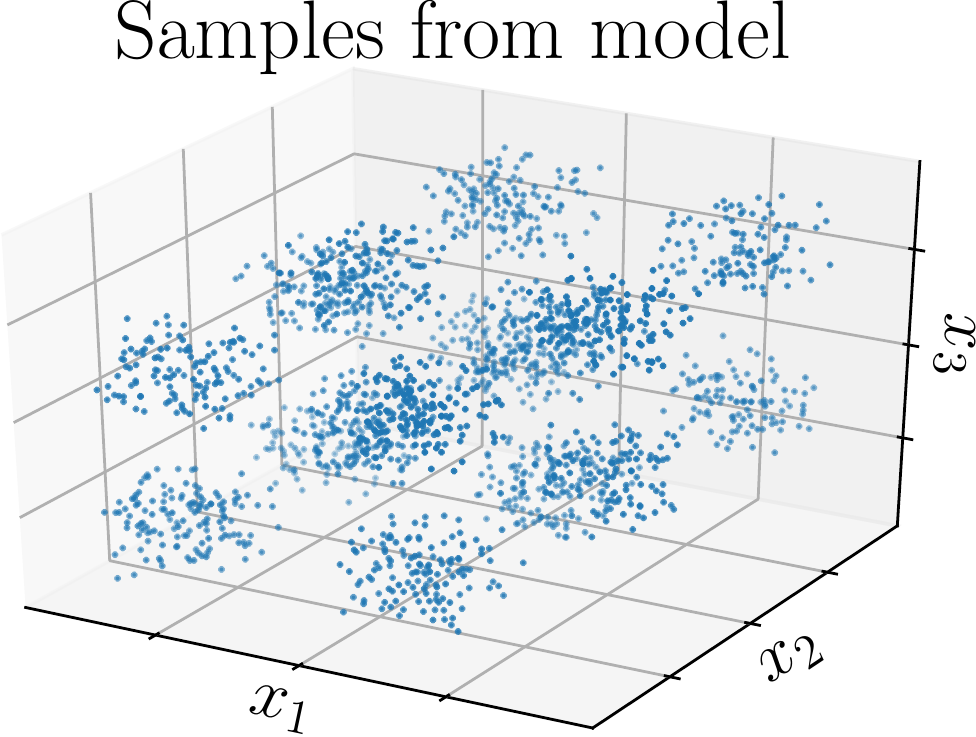}
      \label{}
    \end{subfigure}&
    \begin{subfigure}[c]{\subfigwidth}
      \includegraphics[width=\textwidth]{figs/3d_checkerboard_2d_marg_1_2.pdf}
      \label{}
    \end{subfigure}&
    \begin{subfigure}[c]{\subfigwidth}
      \includegraphics[width=\textwidth]{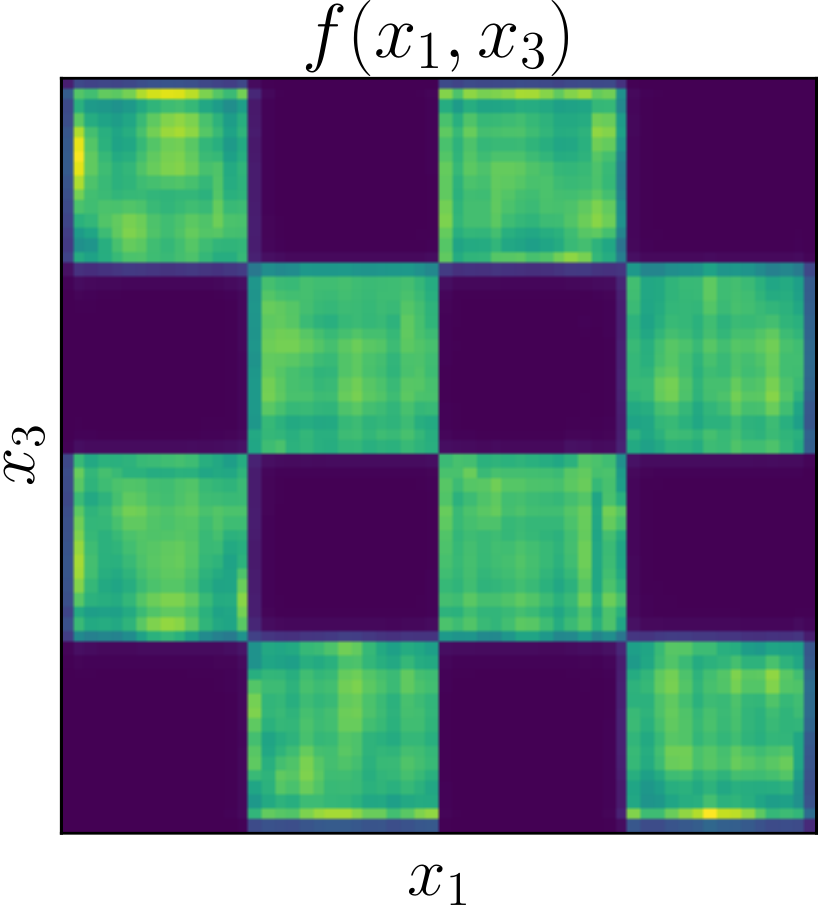}
      \label{}
    \end{subfigure}&
    \begin{subfigure}[c]{\subfigwidth}
      \includegraphics[width=\textwidth]{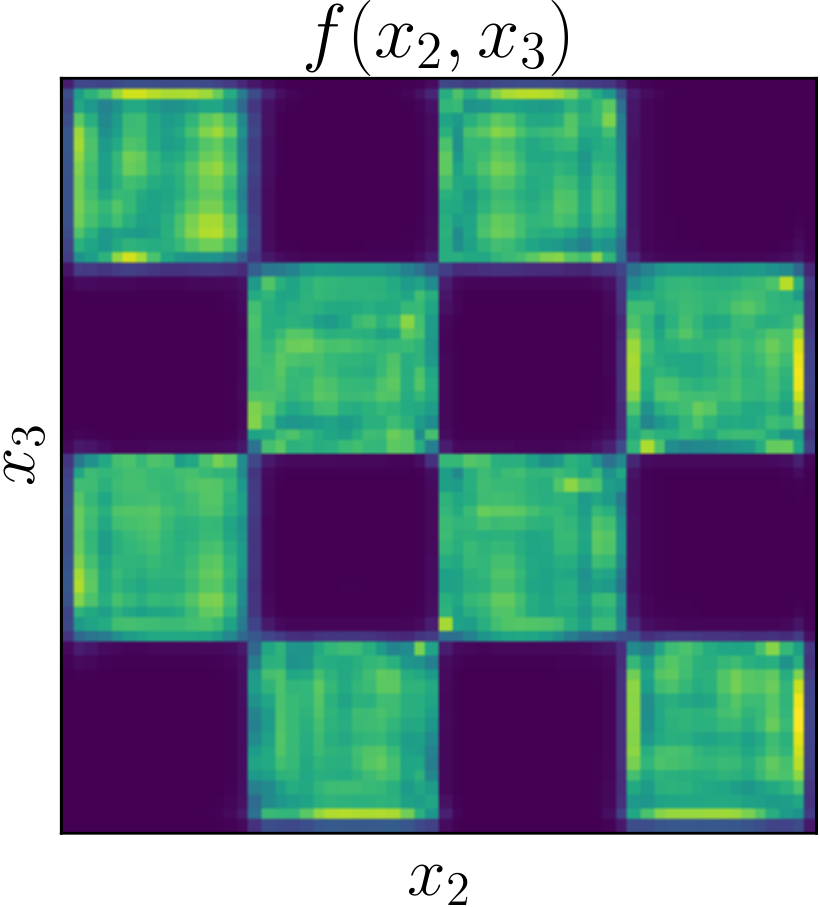}
      \label{}
    \end{subfigure}
    \\[-.1in]
    \begin{subfigure}[c]{\subfigwidth}
    \vspace{.05in}
      \includegraphics[width=\textwidth]{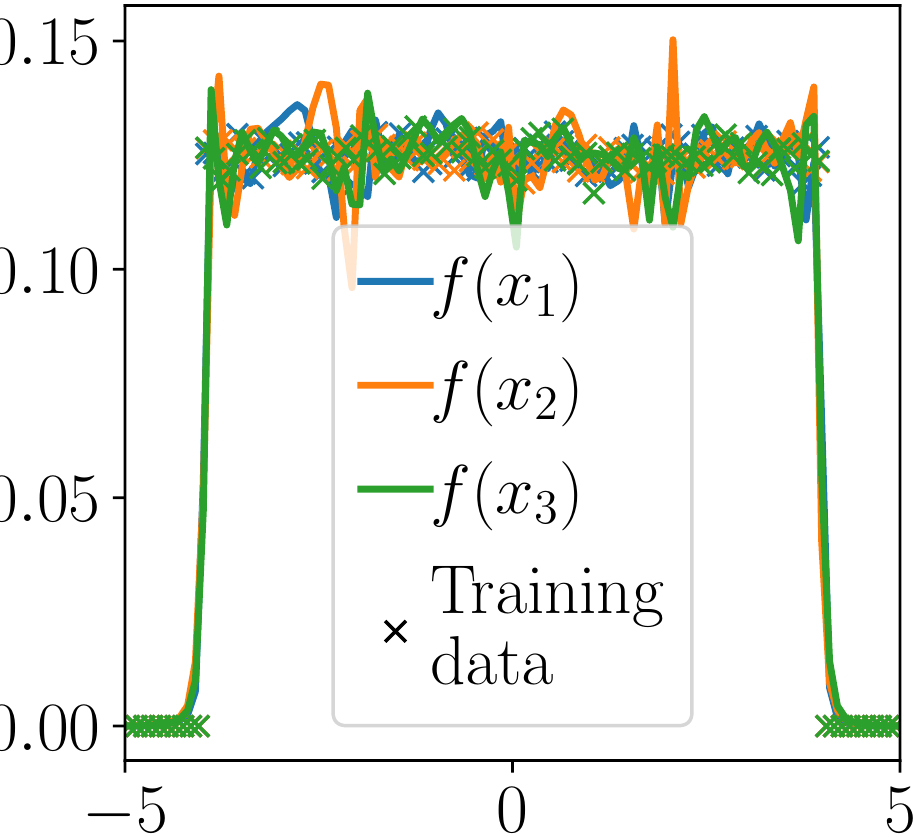}
      \label{}
    \end{subfigure}&
    \begin{subfigure}[c]{\subfigwidth}
      \includegraphics[width=\textwidth]{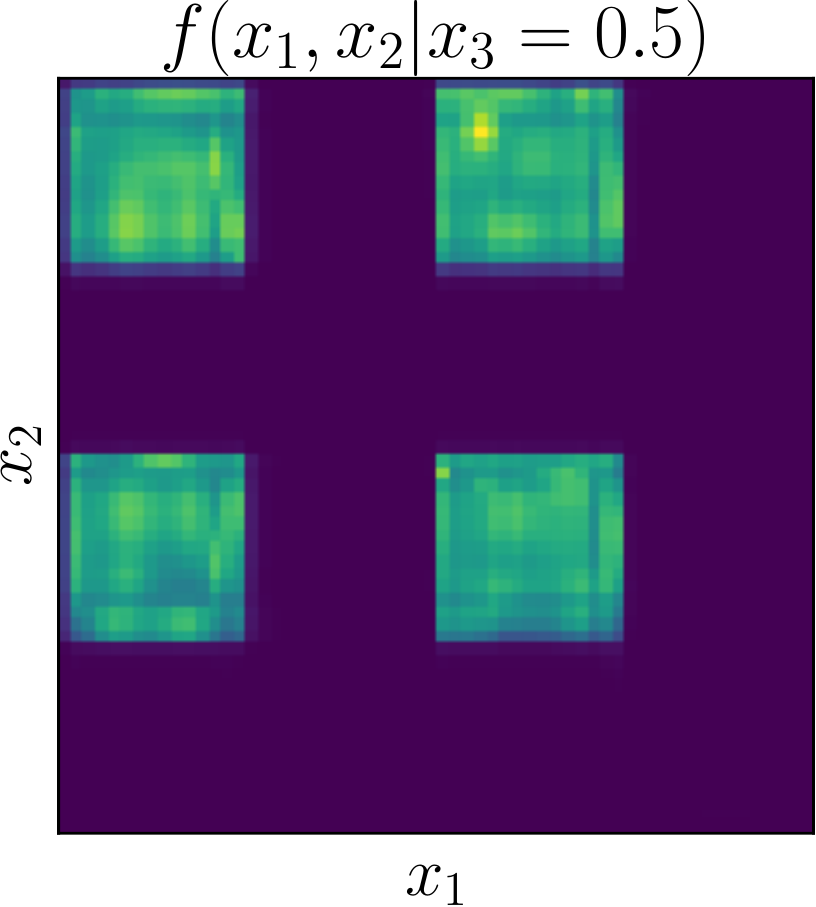}
      \label{}
    \end{subfigure}&
    \begin{subfigure}[c]{\subfigwidth}
      \includegraphics[width=\textwidth]{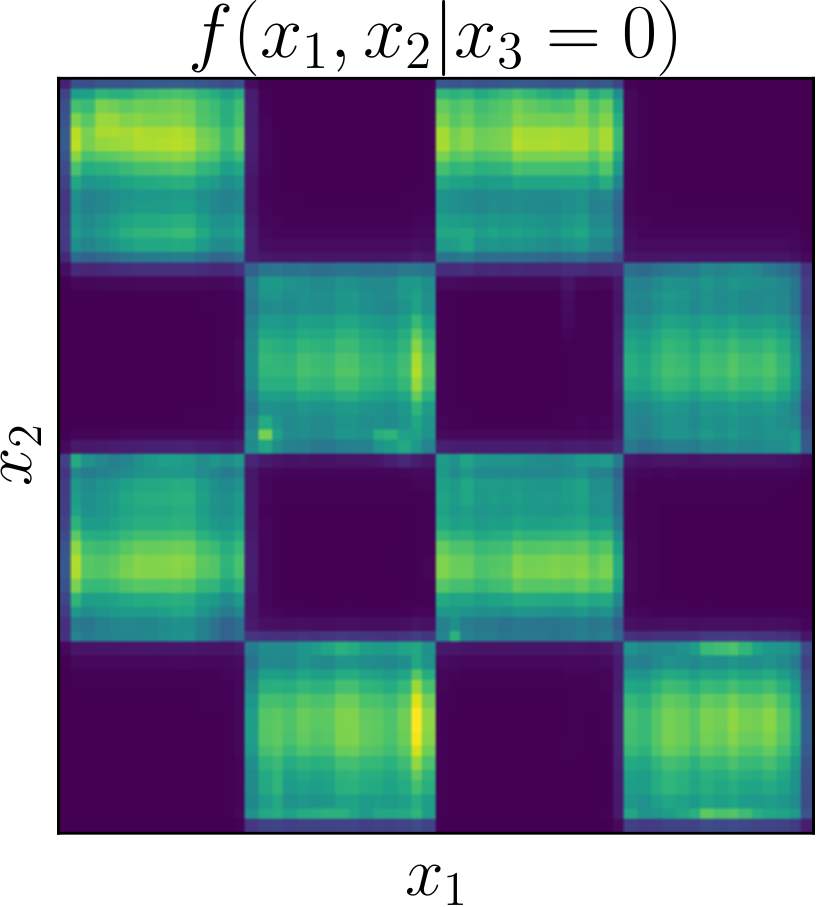}
      \label{}
    \end{subfigure}&
    \begin{subfigure}[c]{\subfigwidth}
      \includegraphics[width=\textwidth]{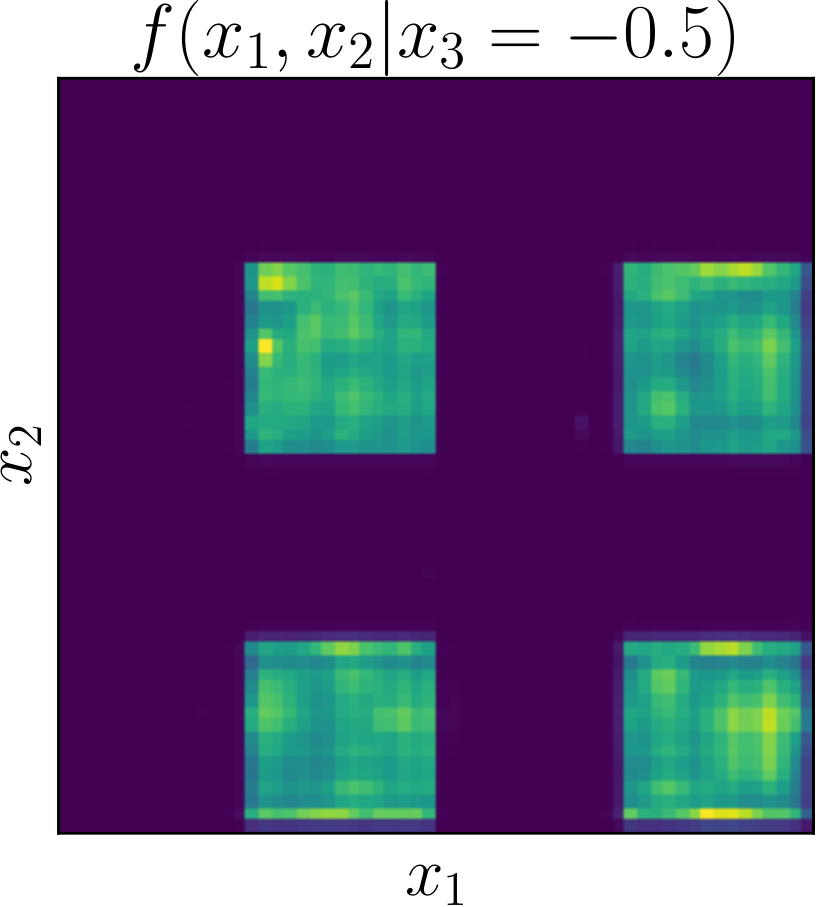}
      \label{}
    \end{subfigure}
  \end{tabular}};
    \draw[black,thick,] (-7,2.3) rectangle (7,6.3);
      \node[text width=3cm] at (0.6,6) 
    {Training data};
\end{tikzpicture}
  \caption{\textbf{Density estimation with closed-form marginals and conditionals.}
  \textit{Top Row:} Samples from a 3D density, and 2D marginal histograms. \textit{Middle Row:} Samples from MDMA after fitting the density, and plots of the learned 2D marginals.
  \textit{Bottom Row:} \textit{Left:} learned 1D  marginals compared to 1D marginal histograms of the training data. \textit{Right:} Learned conditional densities.}
  \label{fig:checkerboard}
\end{figure}

%% file: arxiv/exp_toy_gaussians.tex

\begin{figure}[h]
  \centering
  \renewcommand{\tabcolsep}{1pt}
  \begin{tikzpicture}
    \node {\begin{tabular}[c]{cccc}
\begin{subfigure}[c]{\subfigwidth}
      \includegraphics[width=\textwidth]{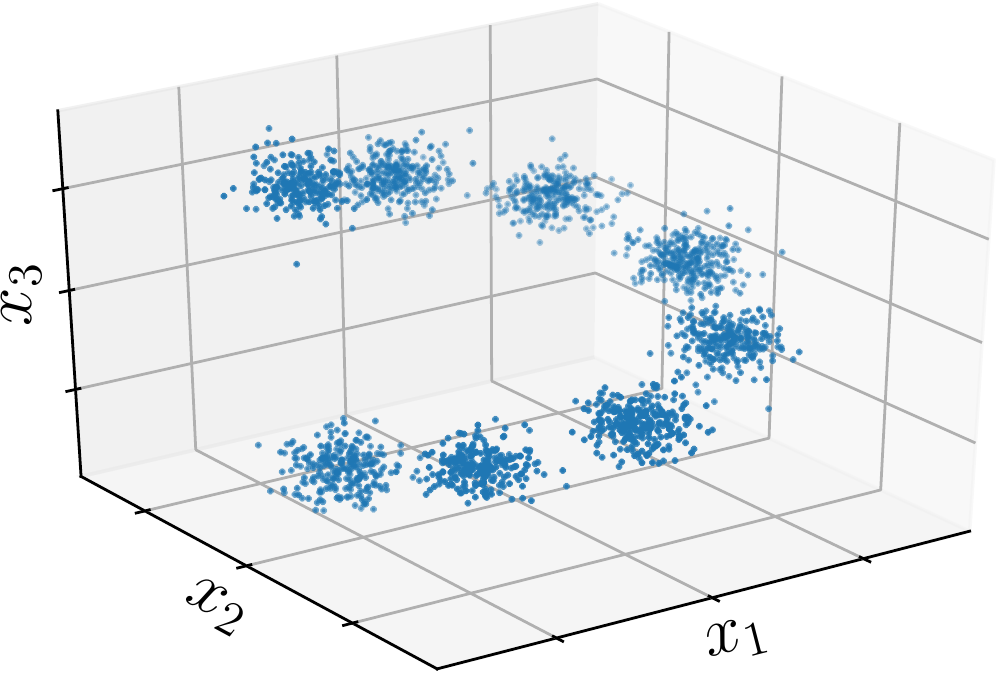}
      \label{}
    \end{subfigure}&
    \begin{subfigure}[c]{\subfigwidth}
      \includegraphics[width=\textwidth]{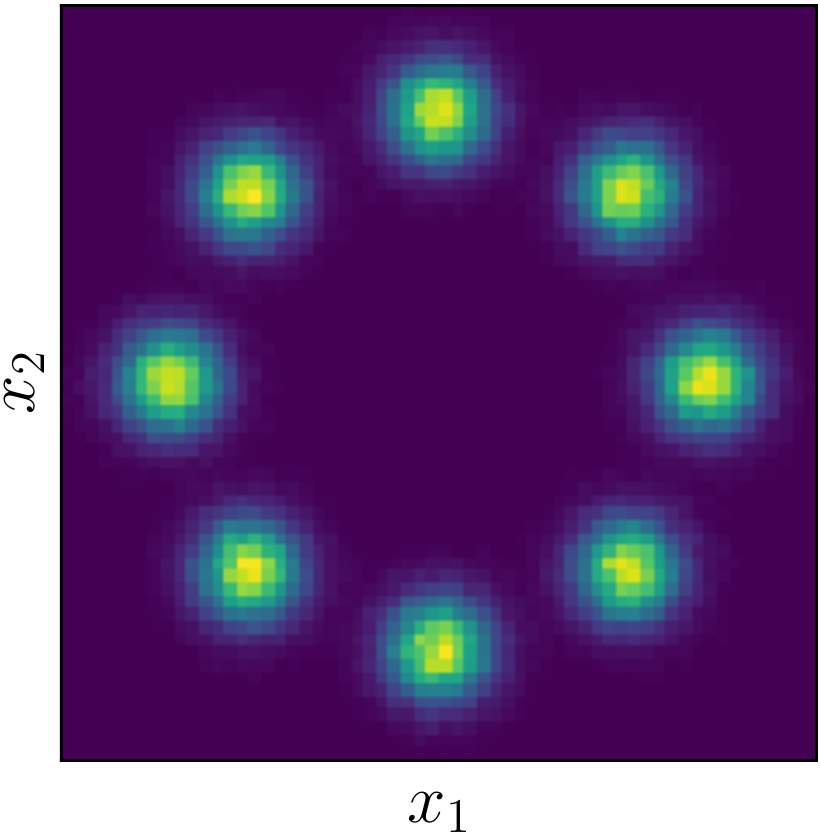}
      \label{}
    \end{subfigure}&
    \begin{subfigure}[c]{\subfigwidth}
      \includegraphics[width=\textwidth]{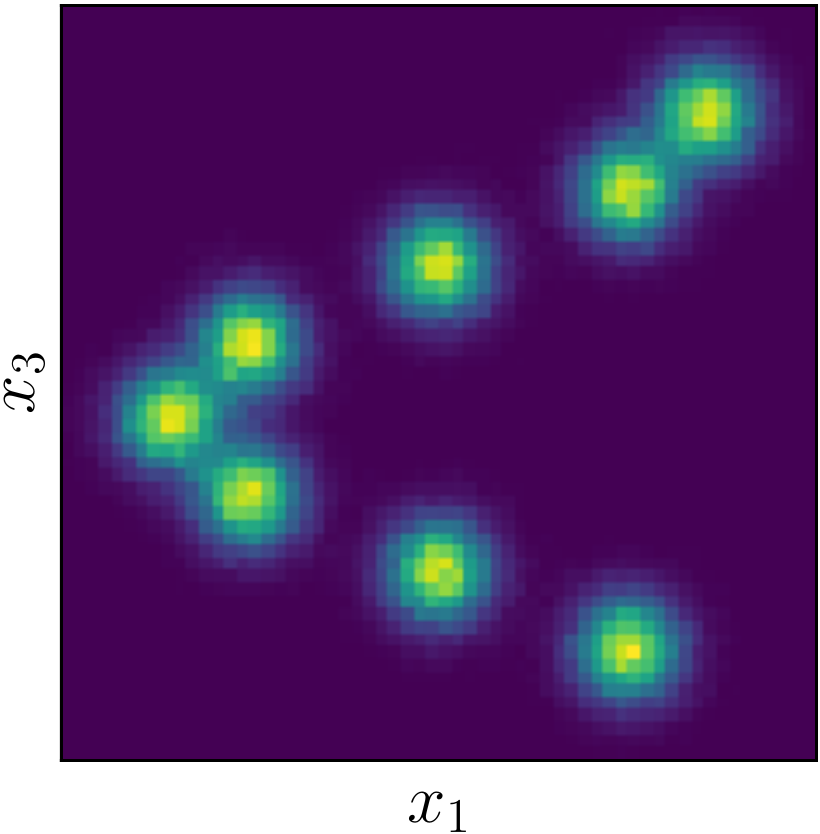}
      \label{}
    \end{subfigure}&
    \begin{subfigure}[c]{\subfigwidth}
      \includegraphics[width=\textwidth]{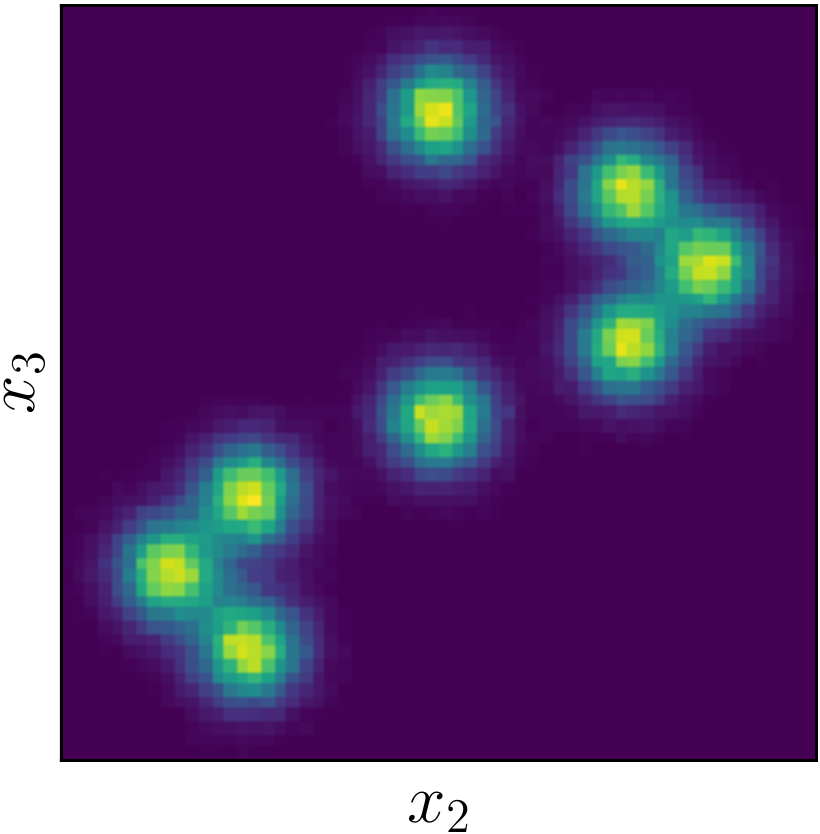}
      \label{}
    \end{subfigure}
    \\[-.1in]
       \begin{subfigure}[c]{\subfigwidth}
      \includegraphics[width=\textwidth]{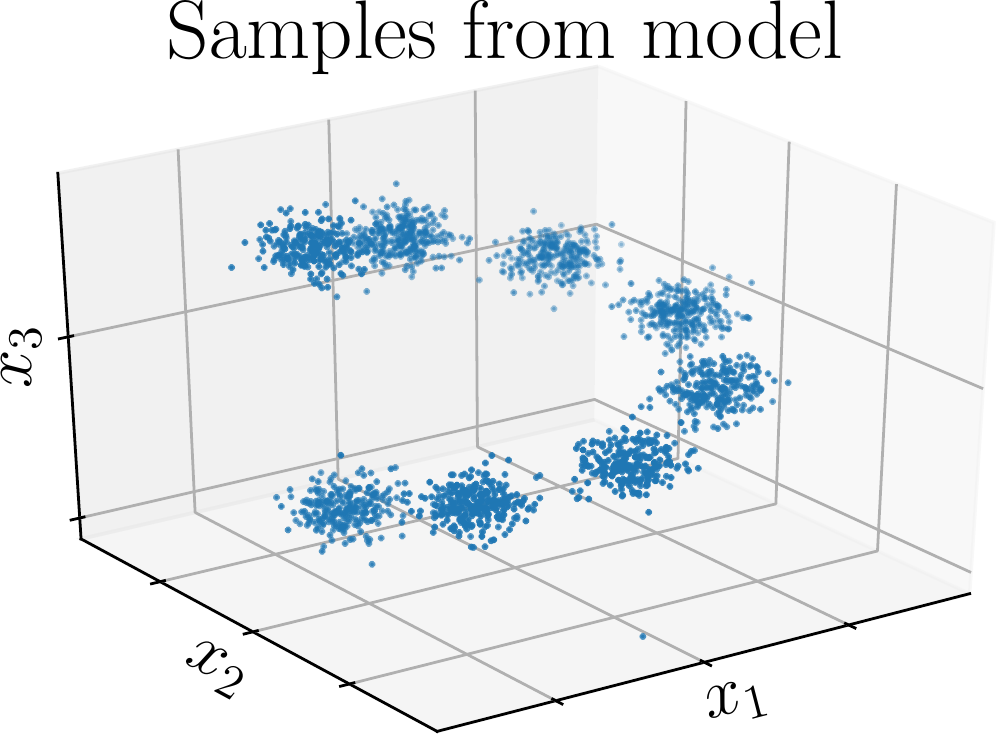}
      \label{}
    \end{subfigure}&
    \begin{subfigure}[c]{\subfigwidth}
      \includegraphics[width=\textwidth]{figs/3d_gaussians_2d_marg_1_2.pdf}
      \label{}
    \end{subfigure}&
    \begin{subfigure}[c]{\subfigwidth}
      \includegraphics[width=\textwidth]{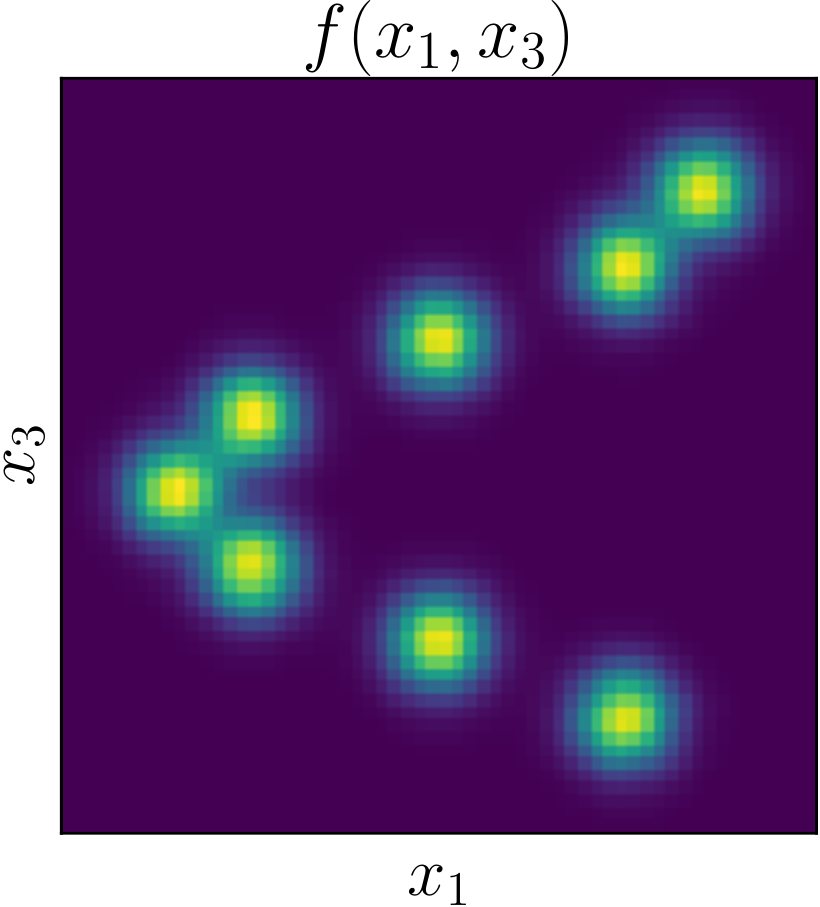}
      \label{}
    \end{subfigure}&
    \begin{subfigure}[c]{\subfigwidth}
      \includegraphics[width=\textwidth]{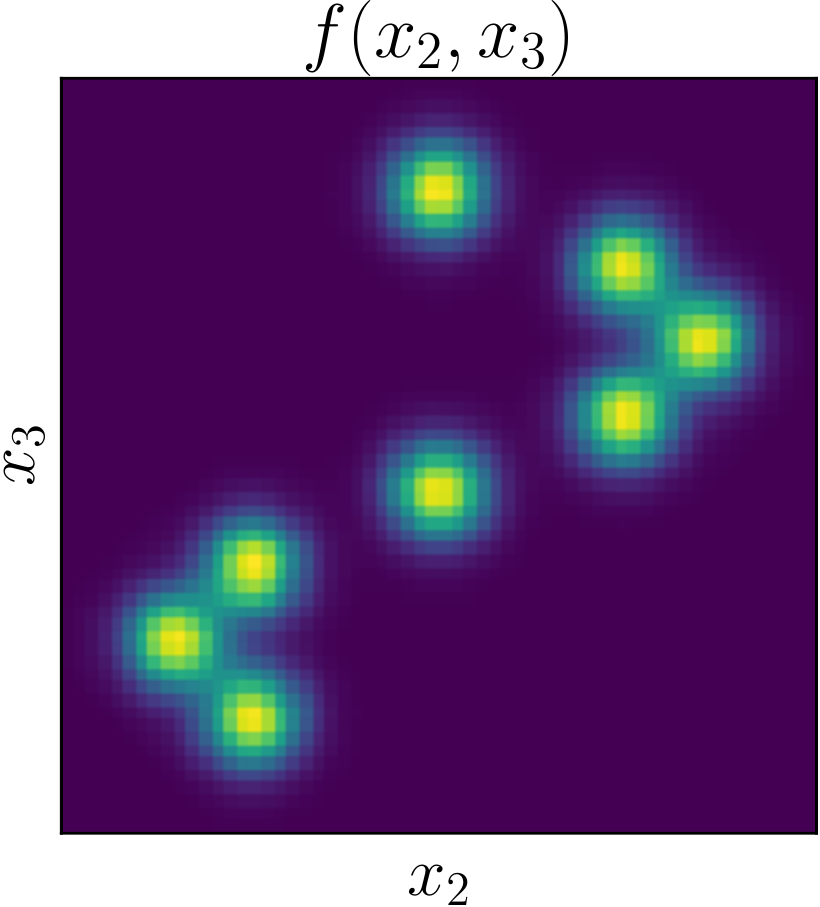}
      \label{}
    \end{subfigure}
    \\[-.1in]
    \begin{subfigure}[c]{\subfigwidth}
    \vspace{.05in}
      \includegraphics[width=\textwidth]{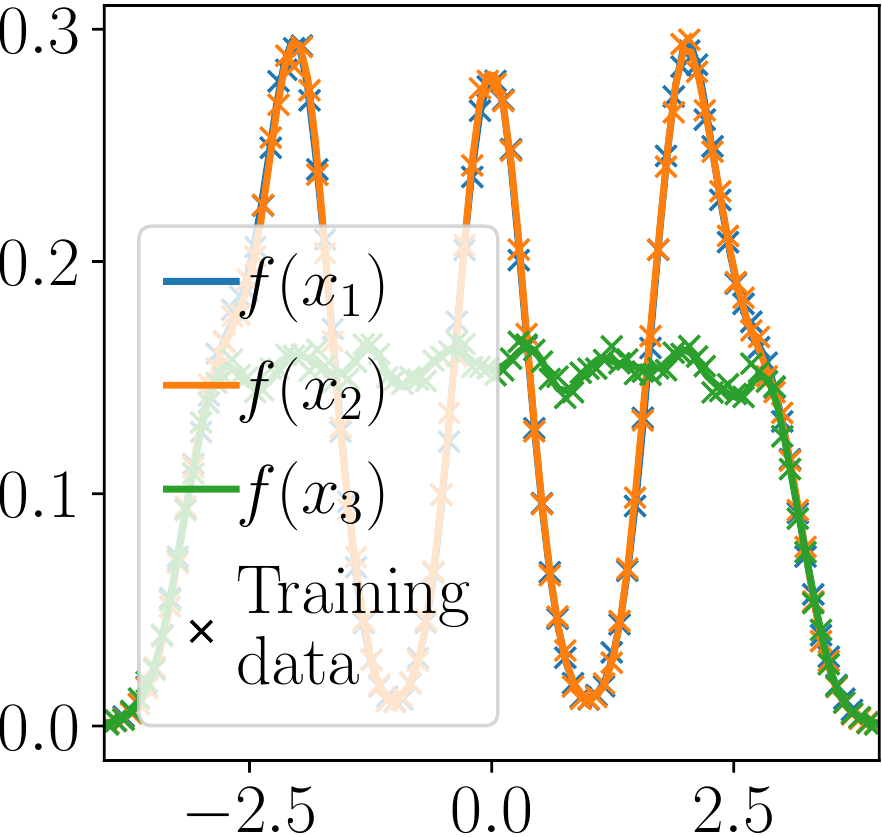}
      \label{}
    \end{subfigure}&
    \begin{subfigure}[c]{\subfigwidth}
      \includegraphics[width=\textwidth]{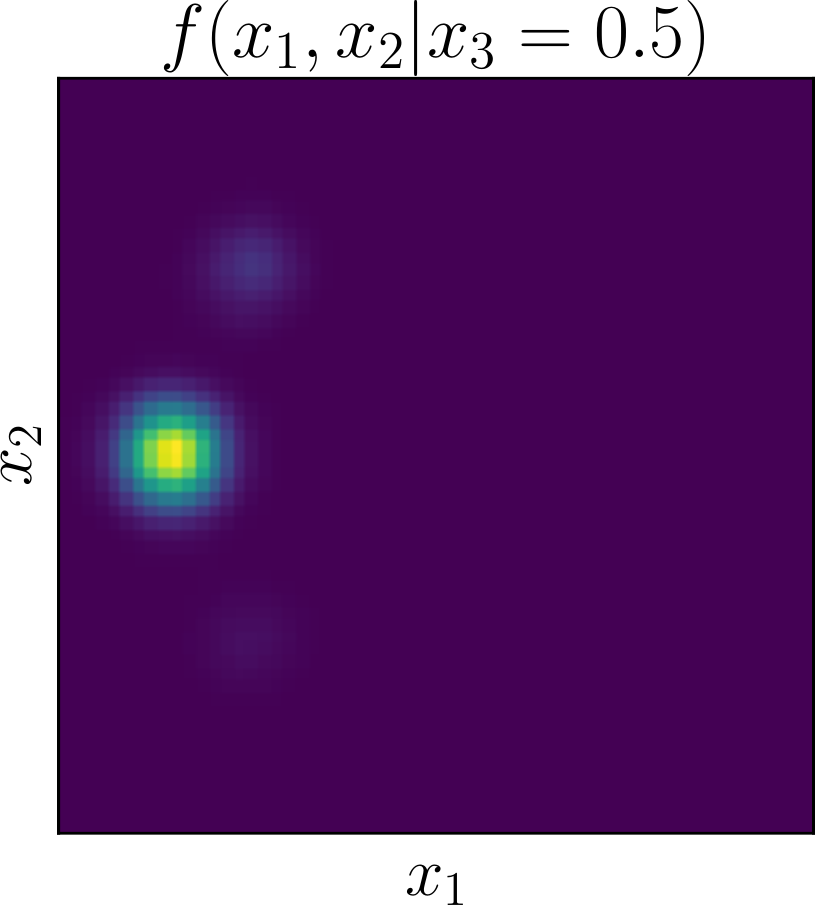}
      \label{}
    \end{subfigure}&
    \begin{subfigure}[c]{\subfigwidth}
      \includegraphics[width=\textwidth]{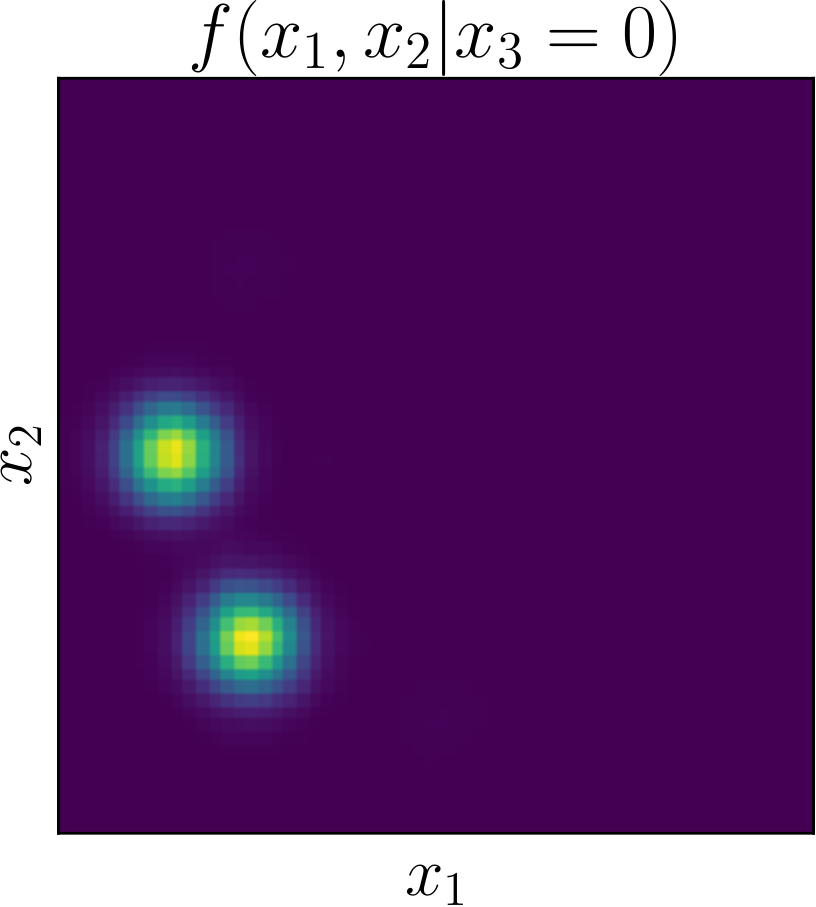}
      \label{}
    \end{subfigure}&
    \begin{subfigure}[c]{\subfigwidth}
      \includegraphics[width=\textwidth]{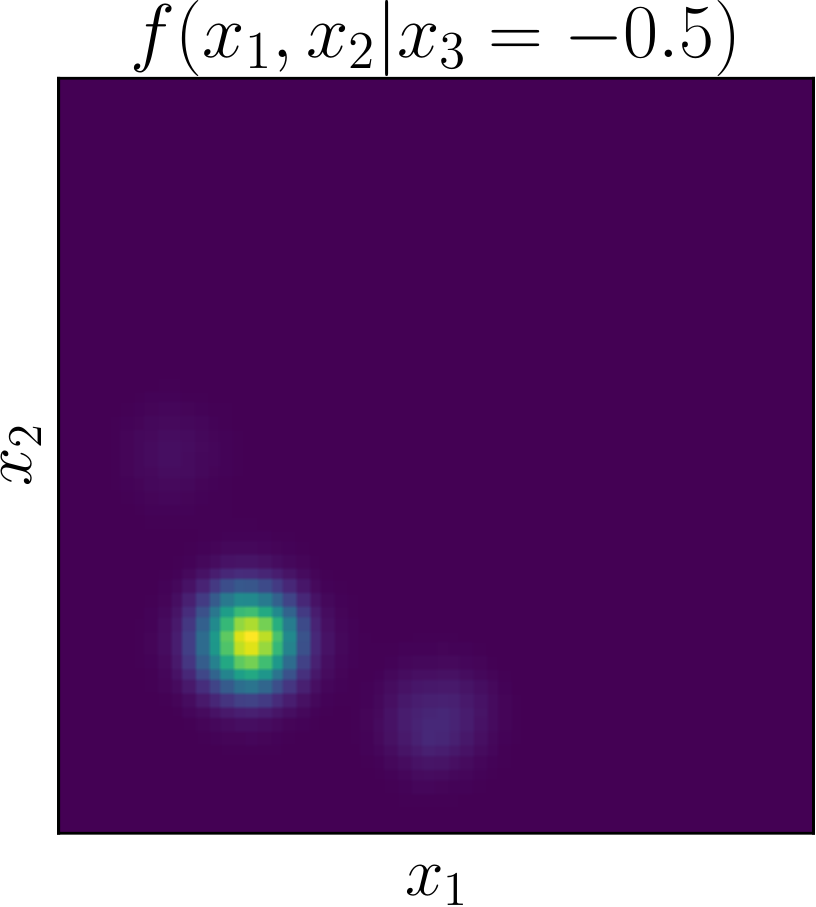}
      \label{}
    \end{subfigure}
  \end{tabular}};
    \draw[black,thick,] (-7,2.3) rectangle (7,6.3);
      \node[text width=3cm] at (0.6,6) 
    {Training data};
\end{tikzpicture}
  \caption{\textbf{Density estimation with closed-form marginals and conditionals.}
  \textit{Top Row:} Samples from a 3D  density, and 2D marginal histograms. \textit{Middle Row:} Samples from MDMA after fitting the density, and plots of the learned 2D  marginal.
  \textit{Bottom Row:} \textit{Left:} learned 1D  marginals compared to 1D marginal histograms of the training data. \textit{Right:} Learned conditional densities.}
  \label{fig:8gaussians}
\end{figure}

%% file: arxiv/design_details.tex

\section{Additional design details} \label{app:design_details}

\subsection{Univariate marginal parameterization} 
\label{app:univariate_marginal_param}

We parameterize the univariate marginal CDF $\varphi(x)$ for some scalar $x \in \mathbb{R}$ using a simple feed-forward network following \cite{balle2018variational}. Recall from section \Cref{sec:theory} that we model the univariate CDFs as functions
\begin{align*}
     \varphi(x) = \mbox{sigmoid} \circ L_{l} \circ \sigma_{l-1} \circ L_{l-1} \circ \sigma_{l-2} \cdots \circ \sigma_1 \circ L_1 \circ \sigma_0 \circ L_0(x),
\end{align*}
where $L_i \colon \R^{n_{i}} \to \R^{n_{i+l}}$ is the affine map $L_i(x) = W_i x + b_i$ for an $n_{i+1} \times n_{i}$ weight matrix $W_i$ with \emph{nonnegative} elements and an $n_{i+1} \times 1$ bias vector $b_i$, with $n_{l+1} = n_0 = 1$ and $n_i = r$ for $i \in [l]$. This is a slightly more general form than the one in \Cref{sec:theory} since we allow the nonlinearities to depend on the layer. For the nonlinearities, we use 
\begin{equation*}
   \sigma_i({x})={x}+{a_i}\odot\tanh({x})
\end{equation*}
for some vector $a_i \in \mathbb{R}^{n_{i+1}}$ with elements constrained to lie in $[-1, 1]$ (the lower bound on $a_i$ is necessary to ensure that $\sigma_i$ are invertible, but the upper bound is not strictly required). This constraint, as well as the non-negativity constraint on the $W_i$, is enforced by setting ${W}_i=\mathrm{softplus}(\tilde{{W}}_i,10),{a_i}=\tanh(\tilde{{a}}_i)$ in terms of some $\tilde{{W}}_{i}\in\mathbb{R}^{n_{i+1}\times n_{i}},\tilde{{a}}_{i}\in\mathbb{R}^{n_{i+1}}$. The softplus function is $\mathrm{softplus}(x,\beta)=\frac{1}{\beta}\log(1+\exp(\beta x))$ and is a smooth, invertible approximation of the ReLU. We typically use small values for $l,r$ in the experiments (see \Cref{app:exp_details}).  

 \subsection{Adaptive variable coupling}

One degree of freedom in constructing a HT decomposition is the choice of partitions of subsets of the variables at every layer over which the products are taken.
This imposes a form of weight-sharing, and it will be natural to share weights between variables that are highly correlated.
As a simple example, let $d=4$ and consider two different HT decompositions
\begin{align*}
   F(x_{1},x_{2},x_{3},x_{4})=\underset{i_{1},i_{2},k}{\sum}\lambda_{k}^{2}\lambda_{k,i_{1},1}^{1}\lambda_{k,i_{2},2}^{1}\varphi_{i_{1},1}(x_{1})\varphi_{i_{1},2}(x_{2})\varphi_{i_{2},3}(x_{3})\varphi_{i_{2},4}(x_{4}),\\
    \widetilde{F}(x_{1},x_{2},x_{3},x_{4})=\underset{i_{1},i_{2},k}{\sum}\lambda_{k}^{2}\lambda_{k,i_{1},1}^{1}\lambda_{k,i_{2},2}^{1}\varphi_{i_{1},1}(x_{1})\varphi_{i_{1},3}(x_{3})\varphi_{i_{2},2}(x_{2})\varphi_{i_{2},4}(x_{4}),
\end{align*}
obtained by coupling $X_1$ in the first layer respectively with $X_2$ and $X_3$.
The univariate marginals for $X_1$, $X_2$ and $X_3$ can then be written as
\begin{align*}
       F(x_{1})=\underset{i_{1},k}{\sum}\lambda_{k}^{2}\lambda_{k,i_{1},1}^{1}\varphi_{i_{1},1}(x_{1}),\   F(x_{2})=\underset{i_{1},k}{\sum}\lambda_{k}^{2}\lambda_{k,i_{1},1}^{1}\varphi_{i_{1},2}(x_{2}),\  F(x_{3})=\underset{i_{2},k}{\sum}\lambda_{k}^{2}\lambda_{k,i_{2},2}^{1}\varphi_{i_{2},3}(x_{3}), \\
       \widetilde{F}(x_{1})=\underset{i_{1},k}{\sum}\lambda_{k}^{2}\lambda_{k,i_{1},1}^{1}\varphi_{i_{1},1}(x_{1}),\ \widetilde{F}(x_{2})=\underset{i_{2},k}{\sum}\lambda_{k}^{2}\lambda_{k,i_{2},2}^{1}\varphi_{i_{2},2}(x_{2}),\ \widetilde{F}(x_{3})=\underset{i_{1},k}{\sum}\lambda_{k}^{2}\lambda_{k,i_{1},1}^{1}\varphi_{i_{1},3}(x_{3}).
\end{align*}
Assume that the variables $X_1$ and $X_2$ are identical. 
In $F$, both of their univariate marginals depend in an identical way on the tensor parameters.
In $\widetilde{F}$ however, additional parameters are required to represent them.
Hence $F$ is a more parsimonious representation of the join distribution.
If $X_1$ and $X_3$ are identical instead, then the converse holds and $\widetilde{F}$ is the more parsimonious representation.
This property extends to any higher-dimensional (e.g., bivariate) marginals.

In data with spatial or temporal structure (e.g. if the variables are image pixels) there is a natural way to couple variables based on locality. When this is not present, we can adaptively construct the couplings based on the correlations in the data using a simple greedy algorithm. After constructing an empirical covariance matrix from a minibatch of data, we couple the two variables that are most correlated and have not yet been paired. We repeat this until we couple all the groups of variables. Then we "coarse-grain" by averaging over blocks of the covariance matrix arranged according to the generated coupling and repeat the process, this time coupling subsets of variables. We find that this coupling scheme improves performance compared to naive coupling that does not take correlations into account. 

\subsection{Initialization}

As in the univariate case, the non-negativity constraint of the HT tensor parameters $\lambda_{k,k',j}^{i}$ is enforced by defining $\lambda_{k,k',j}^{i}=\mathrm{softplus}\left(\tilde{\lambda}_{k,k',j}^{i},20\right)$ for some $\tilde{\lambda}_{k,k',j}^{i}\in\mathbb{R}$. 

As is standard, we initialize independently the elements of the univariate PDF weights $\tilde{W}_i$ as zero-mean gaussians with variance $1/n_{\mathrm{fan_in}}$, the $\tilde{a}_i$ as standard gaussians and the $b_i$ as 0. The initialization of the HT parameters is $\tilde{\lambda}_{k,k',j}^{i} = m * \delta_{k,k'}$ for $1\leq i\leq p-1$. This initialization is chosen so that after applying the softplus the matrix $\lambda_{\cdot,\cdot,j}^{i}$ is close to an identity at initialization, which we have found facilitates training compared to using a random initialization. Benefits of such ``orthogonal'' initialization schemes have also been shown for deep convolutional networks \cite{xiao2018dynamical, Blumenfeld2020-lu}. The final layer $\tilde{\lambda}_{k}^{p}$ are initialized as zero mean gaussians with variance $0.3/m$.  

\subsection{From HT to MERA}

The choice of the diagonal HT decomposition \cref{eq:HT} is convenient, yet there is a wealth of other tensor decompositions that can be explored. Here we highlight one such decomposition that generalizes the diagonal HT and could potentially lead to more expressive models. It is based on \cite{Giovannetti2008-pk}. 

Let $F^{\mathrm{M},1}\colon \R^d \to \R^{m \times d}$ be a matrix-valued function such that $F^{\mathrm{M},1}_{i,j}(\bm x) = F^{\mathrm{HT},1}_{i,j}(\bm x) = \varphi_{i,j}(x_j)$.
For $l \in \{2, \dots, \log_2 d\}$, define the matrix-value functions $F^{\mathrm{M},1}\colon \R^d \to \R^{m \times d/2^{l-1}}$ recursively by 
\begin{align*}
F^{\mathrm{M},l}_{i,j}(\bm x)  &= 
  \sum\limits_{k}^{m}  \lambda_{k,i,j}^{l-1} \{ \chi_{1,k,i,j}^{l-1}\varphi_{k,2j-1}^{\mathrm{M},l-1}(\bm x)\varphi_{k,2j}^{\mathrm{M},l-1}(\bm x) +  \chi_{2,k,i,j}^{l-1}\varphi_{k,2j-1}^{\mathrm{M},l-1}(\bm x)\varphi_{k + 1,2j}^{\mathrm{M},l-1}(\bm x)\}
\end{align*}
with $k+1 \equiv 1$ when $k = m$, $\lambda^{l}$ as in \eqref{eq:HT_recurs}, and $\chi^{l}$ a $2 \times m \times m \times d/2^{l}$ tensor with nonegative elements satisfying $\chi_{1,k,i,j}^{l}+\chi_{2,k,i,j}^{l} = 1$.
The MERA parametrization of a distribution can then be written as
\begin{align*}
 F_{\mathrm{M}}(\bm x) &= \sum\limits_{k=1}^m a_{k} \varphi_{k,1}^{\mathrm{M},\log_{2}d}(\bm x)\varphi_{k,2}^{\mathrm{M},\log_{2}d}(\bm x),
\end{align*}
with $a \in \mathbb{R}^m$
Since the conditions on $\lambda^{l}$ and $\chi^{l}$ imply $\sum\limits_{k=1}^m\lambda_{k,i,j}^{l}(\chi_{1,k,i,j}^{l}+\chi_{2,k,i,j}^{l}) = 1$, this parametrization clearly results in a valid CDF.
Note that $\chi_{1,k,i,j}^{l}+\chi_{2,k,i,j}^{l} = 1$ leads to $\chi^{l}$ having only $m \times m \times d/2^{l}$ free parameters.
For $d=4$, we have
\begin{align*}
  F_{\mathrm{M}}(\bm x) 
  &= \sum\limits_{k_1,k_2,k_2'=1}^m a_{k_1} \lambda_{k_2,k_1,1}^{1} \lambda_{k_2',k_1,2}^{1}\sum\limits_{i_1,i_2=1}^2  \chi_{i_1,k_2,k_1,1}^{1}\chi_{i_2,k_2',k_1,2}^{1}  \\
  &\phantom{ = \sum\limits_{k_1,k_2,k_2',k_3,k_3',k_3'',k_3'''=1}^m } \times \varphi_{k_2,1}^{\mathrm{M},1}(\bm x)\varphi_{k_2 + i_1 - 1,2}^{\mathrm{M},1}(\bm x)\varphi_{k_2',3}^{\mathrm{M},1}(\bm x)\varphi_{k_2' + i_2 - 1,4}^{\mathrm{M},1}(\bm x).
  \end{align*}